\documentclass[shortpaper]{clv2025}

\jvol{vv}
\jnum{nn}
\jyear{2026}

\usepackage{amsmath}
\usepackage{booktabs}

\usepackage{graphicx}
\usepackage{multirow}
\usepackage{makecell}
\usepackage{adjustbox}
\usepackage{float}

\usepackage{xspace}
\NewDocumentCommand{\tbd}{o}{\textcolor{red}{\IfNoValueTF{#1}{??}{#1}}\xspace}
\NewDocumentCommand{\tbdref}{o}{\colorbox{orange}{[ref]}}
\NewDocumentCommand{\tbdfig}{m o}{\colorbox{magenta}{[fig #1]}}

\usepackage{xcolor}
\definecolor{mygreen}{HTML}{39843B}
\definecolor{myblue}{HTML}{2652B0}

\usepackage{hyperref}
\usepackage{cleveref}
\usepackage{xr-hyper}

\dochead{Short Paper} 

\pageonefooter{Action editor: Keisuke Sakaguchi. Submission received: 26 January 2026; revised version received: 29 June 2026; accepted for publication: 24 July 2026.}

\runningtitle{Eye Tracking Based Cognitive Evaluation of ARA Methods}
\runningauthor{Gruteke Klein et al.}

\begin{document}

\title{Eye Tracking Based Cognitive Evaluation of Automatic Readability Assessment Methods}

\author{Keren Gruteke Klein\thanks{Corresponding author}, Shachar Frenkel, Omer Shubi, Yevgeni Berzak}

\affilblock{
    \affil{Faculty of Data and Decision Sciences, Technion – Israel Institute of Technology, Haifa, Israel\\\quad \email{gkeren@campus.technion.ac.il}}
}

\maketitle

\begin{abstract} 
Automatic methods for scoring text readability have been studied for over a century, and are widely used in research and in user-facing applications in many domains. Thus far, the development and evaluation of such methods have primarily relied on two types of offline human behavioral data, performance on reading comprehension tests and ratings of text readability levels. In this work, we instead focus on a fundamental and understudied aspect of readability, real-time reading ease, captured with online reading measures using eye tracking. We introduce a new cognitive evaluation framework for readability scoring methods that quantifies their ability to account for reading ease, while controlling for content variation across texts. Applying this evaluation to prominent traditional readability formulas, NLP-based methods, commercial systems used in education, and frontier LLMs suggests that they are all poor predictors of English reading ease in adults as compared to word properties commonly used in psycholinguistics for the prediction of reading times. This outcome holds across L1 and L2 speakers, different reading regimes, and textual units of different lengths. Our results reveal an important limitation of a wide range of methods for readability scoring, highlight the utility of real-time behavioral benchmarks for readability research, and call for new, cognitively driven readability scoring approaches that can better account for how humans experience texts in real time.
\end{abstract}


\clearpage

\section{Introduction}

Over more than a century, researchers have been developing methods for automated scoring of linguistic readability of texts, a research area often referred to as Automatic Readability Assessment (ARA). ARA has been flourishing due to its societal importance in domains such as education, health care, law and media.
From its early days, text readability assessment has been of central interest in psychology and education research \citep{kintsch-vipod,dubay2004}. In the past few decades, it has gained considerable traction in Natural Language Processing (NLP), where feature extraction tools and machine learning driven systems for readability scoring and especially readability classification of passages \citep[e.g.,][]{heilman2007combining,pitler2008revisiting,kate2010learning,vajjala2012improving,filighera2019automatic,vajjala-2018-onestopenglish,mohammadi2019text,meng2020readnet,arase-etal-2022-cefr} and sentences \citep[e.g.,][]{vstajner2017automatic,brunato2018sentence,liu2025automatic} have been very active research areas (see \citet{vajjala2022trends} for a recent overview). With the recent rise of user-facing Large Language Models (LLMs), there has also been a growing interest in harnessing their impressive capabilities for readability scoring \citep{trott-riviere-2024-measuring,farajidizaji-etal-2024-possible, creutz-2024-correcting}.

While the concept of readability may seem intuitive, or as \citet{kintsch-vipod} put it, ``We know (in our hearts) what we mean by readability'', it turned out to be challenging to define which theoretical constructs are key to linguistic readability and perhaps even more so, what its behavioral indexes are. One construct that received much attention in readability research is \emph{comprehension ease} \citep{klare1963measurement}. It is typically operationalized by offline comprehension measures: querying text comprehension via reading comprehension tasks. Higher comprehension scores are then taken to indicate higher text readability. Many readability formulas developed in the 20th century took this approach by regressing text properties like sentence length, number of syllables per word, and number of complex words on reading comprehension scores \cite[among others]{flesch1948new,dale1948formula,flesch-kincaid1975}. Some of these formulas are still commonly used today. 

A common alternative methodology is \emph{human ratings}, where readers are asked to judge the readability level of a text according to a given annotation scheme. This approach often takes a more holistic stance towards readability, rather than targeting a specific readability construct. Human annotations of readability became widespread with the rise of modern NLP in the 21st century, in which reliance on human annotations for training text processing systems has been standard practice \citep{vajjala2022trends}. Following the recent dramatic advances in NLP with the introduction of LLMs, researchers have also been exploring prompting off-the-shelf LLMs to provide such annotations \citep{trott-riviere-2024-measuring,farajidizaji-etal-2024-possible, creutz-2024-correcting}. 

Although reading comprehension and annotation-based approaches have considerably advanced the study of readability, they capture \emph{offline} behavioral signals that do not directly speak to a fundamental aspect of a readable text, namely the degree of \emph{reading ease} while interacting with the text in \emph{real time}. Reading ease was proposed early on as a key aspect of readability \citep[e.g.,][]{dale-chall1949,kintsch-vipod}. 
For example, Rudolf Flesch writes in the context of the Flesch Reading Ease score ``For many obvious reasons, the grade level of children answering test questions is not the best criterion for general readability. Data about the ease and interest with which adults will read selected passages would be far better. But such data were not available at the time the first formula was developed, and they are still unavailable today'' \citep{flesch1948new}.

However, only a handful of studies have operationalized readability using behavioral signals from reading \citep{Mcclusky1934AQA,clare1957,nahatame2021text,hollenstein2022patterns,nishikawa2013pilot,scozzaro2024reform}. \citet{singh2016quantifying}, \citet{gonzalez2018learning} and \citet{gooding2021predicting} used behavioral reading ease data as features for improving the performance of text readability classification into human generated or human labeled readability levels, but crucially, not for system evaluation. Predicted reading times were also recently leveraged as a readability signal for steering text generation \cite{sauberli2026controlling}.
Throughout the 20th and 21st centuries, the dominant paradigms remained comprehension and rating based, in part due to the scarcity of behavioral data from the reading process. 
The lack of focus on reading ease based benchmarking in ARA is especially apparent when compared to research on typographic determinants of readability, such as font characteristics, and word and character spacing \citep{beier2022readability}. In this area of research, reading speed, often measured in words per minute (WPM), is a standard evaluation measure for text readability \citep{wallace2022towards,wallace2022space,day2024influence}.

Although behavioral traces of real-time language processing have not received much attention in ARA, they have been studied extensively in the psychology of reading and psycholinguistics. In these areas of research, multiple behavioral methodologies are commonly used for capturing language comprehension processes as reading unfolds over time, including self-paced reading \citep{aaronson1976performance,mitchell1978effects}, Maze \citep{forster2009maze,boyce2020maze}, and mouse tracking while reading \citep{wilcox2024mouse}. Perhaps the most informative among these methodologies is eye tracking during reading, which allows capturing real-time behavioral traces of comprehension processes in very fine detail \citep{rayner1998,rayner2006eye,schotter2025beginner}.

Furthermore, largely independently from readability research, eye tracking and other behavioral methodologies were used in psycholinguistics to study a construct that is closely related to reading ease, namely \emph{processing difficulty}. Over the past few decades, a large body of work has developed theoretical frameworks that aim to account for processing difficulty in language comprehension \cite[e.g.,][]{levy2008,gibson1998linguistic}, as well as text-based measures of processing difficulty that can be derived from these frameworks.  Theoretically motivated textual predictors of processing difficulty have been used extensively in psycholinguistics, and have often been evaluated using eye tracking and other behavioral measures \cite[][among others]{kliegl2004length,shain2024large,shain2024word,rayner2004effects}. 
In ARA, on the other hand, psycholinguistic text measures have received relatively little attention either as stand-alone or add-on readability measures. A few notable exceptions are \citet{howcroft-demberg2017}, who predicted text readability levels using psycholinguistic measures, and \citet{martinc2021supervised}, who developed perplexity-derived measures for unsupervised readability scoring, further evaluated in \citet{ehara2021evaluation}.  \citet{miaschi2020neural} examined the correlation of perplexity with ARA scores derived from lexical, morphological, and syntactic features.  \citet{xia2019text} used perplexity as one of the features in a supervised readability system.  Despite these advancements, the use of psycholinguistic measures in readability research remains limited. 

In this work, we bring psycholinguistic research closer to the study of readability by introducing a \emph{cognitive} framework for evaluating readability scoring methods. This framework focuses on real-time \emph{reading ease} and captures it via behavioral traces of eye movements in reading, which are known to index online processing difficulty. The introduction and application of this framework in the current study reflect several advancements compared to prior work, which used eye tracking in the context of readability \citep{Mcclusky1934AQA,clare1957,nahatame2021text,hollenstein2022patterns,nishikawa2013pilot,scozzaro2024reform}. 

First, we argue for and introduce the use of eye tracking data for the \emph{evaluation of readability scoring methods}. The presented evaluation method provides readability scores on a continuous scale, and addresses a key methodological challenge in readability research, which concerns the entanglement of text difficulty aspects stemming from topical variation across texts, and those related to linguistic complexity \citep{vajjala2022trends}. Our approach decouples the two by leveraging manual text simplifications, thus allowing focusing on \emph{writing style}, which is central to many definitions of readability \citep{kintsch-vipod,dubay2004,vajjala2022trends}.

Second, we carry out \emph{comprehensive evaluations} for a variety of prominent readability scoring methods, including traditional readability formulas, modern NLP-based methods, commercial systems used in education and frontier LLMs. These methods are benchmarked against theoretically and empirically motivated textual predictors of processing difficulty from the psycholinguistic literature, including idea density \citep{kintsch1972notes}, integration cost \citep{gibson1998linguistic,gibson2000dependency}, uniform information density \citep{jaeger2006speakers}, embedding depth \citep{van2012connectionist}, word entropy, surprisal \citep{hale2001,levy2008}, frequency and length. Our study offers \emph{large-scale} analyses across L1 and L2 readers, different reading regimes, and different textual units.

Our evaluations yield a consequential result. Despite the extensive efforts invested over decades in automated readability scoring methods, as well as the unprecedented linguistic and meta-linguistic capabilities of current LLMs, in adult readers of English they all have weak predictive power for reading facilitation in simplification, and by extension reading ease that is associated with writing style. While some of the theoretically based psycholinguistic measures perform similarly poorly, in most cases, standard word properties from the psycholinguistic literature that are known to be robust predictors of reading times perform well on our evaluations. These measures are word entropy, word length, word frequency and surprisal: the negative log probability of a word in context. Surprisal tends to be the most predictive index of reading ease overall. 

These results hold across different adult reader groups (L1 and L2), reading regimes (reading for comprehension and information seeking) and textual units (passages and sentences). They provide, for the first time, a comprehensive picture of the limited ability of ARA methods developed over nearly a century to account for a key aspect of how readers experience texts in real time. Importantly, they warrant more caution in the use of existing readability assessment methods and LLMs for readability scoring, especially in high-stakes scenarios. Furthermore, they call for new, more cognitively driven and empirically validated methodologies for readability scoring.

\section{Methods}
\label{sec:methods}

\subsection{Eye Movement Data} 

\subsubsection{Overview}

Our study relies on a combination of two datasets, OneStop \texttt{v1.0.3} \citep{onestop2025}, which is a publicly available dataset, and OneStopL2 \texttt{v0.1}, a dataset that was collected as part of this work. Both are broad-coverage datasets of eye movements in English reading with identical textual materials and an identical experimental design, collected with an EyeLink 1000 Plus eye tracker (SR Research) at a sampling rate of 1000 Hz. The combined dataset, which we refer to as OneStopL1\&L2, contains over 3.7 million word tokens for which eye movements were collected during first reading, from 638 adult participants.
Crucially, both OneStop and OneStopL2 use a parallel corpus of texts in their original and human-simplified forms, which makes them uniquely suited for our study.

\subsubsection{Textual materials} OneStopL1\&L2 uses textual materials from the OneStopQA dataset \citep{starc2020,vajjala-2018-onestopenglish}. They consist of 30 Guardian news articles with 4--7 paragraphs (162 paragraphs in total) from the News Lessons section of the English language-learning portal onestopenglish.com by Macmillan Education. Articles were simplified from their original ``Advanced'' version to a simpler ``Elementary'' version by experienced language teachers who are staff members of onestopenglish.com. The simplification method is ``intuitive'', and relies on experience and subjective judgment rather than on structured guidelines and metrics \citep{simensen1987adapted,young1999linguistic}. Each paragraph has three manually composed multiple-choice reading comprehension questions. 
Each question can be answered based on either of the paragraph's difficulty level versions. To support the current study, we manually created an additional sentence-level segmentation and alignment between the two versions of each paragraph. This alignment enables analyses not only at the passage level, but also for individual sentences. We have further manually annotated the aligned sentence pairs with the types of the performed edits according to the simplification edit type taxonomy of \citet{xu-etal-2015-problems}. Additional details on the annotation procedure and annotation agreement rates are provided in Appendix~\ref{sec:additional-datasets-simp-types}.

Table~\ref{table:text-stats} presents summary statistics of the OneStop textual materials across the two text difficulty levels. Appendix~\ref{sec:additional-datasets-app} \Cref{table:text-stats-app} presents these statistics along with statistics from two other commonly used corpora in the simplification and readability literature, PWKP \citep{zhu-etal-2010-monolingual} and CLEAR \citep{crossley2021commonlit}. 
Appendix~\ref{sec:additional-datasets-simp-types} \Cref{fig:simplification_types_comparison} further presents simplification type statistics for OneStop, compared with two other simplification datasets, PWKP and Newsela \citep{xu-etal-2015-problems}. 

\begin{table}[ht]
\caption{Statistics of the original ``Advanced'' and simplified ``Elementary'' versions of OneStopQA texts. Word length excludes punctuation. Frequency counts are from Wordfreq \citep{robyn_speer_2018}. Surprisal is $-\log_2(p(word|context))$, where \textit{context} denotes the paragraph words preceding the current \textit{word}, from the Pythia 70M language model \citep{biderman2023pythia}.
Mean values are reported with 95\% confidence intervals. The rightmost column presents the significance of a t-test comparing the means of the original and simplified text versions: ns ($p \geq 0.05$), *** $(p < 0.001)$.}

\centering
\begin{tabular}{@{}llll@{}}
\toprule
      & \textbf{Original} & \textbf{Simplified} & \textbf{p} \\ \midrule
Number of paragraphs  &  162 & 162 & NA \\ 
Number of sentences  &  936 & 931 & NA \\
Number of questions  &  486 & 486 & NA \\ \midrule
Words per paragraph  &  119.92 ± 4.33 & 97.14 ± 3.66 & *** \\
Sentences per paragraph  &  5.78 ± 0.31 & 5.75 ± 0.27 & ns \\
Sentence length (words)  &  20.76 ± 0.66 & 16.90 ± 0.51 & *** \\ \midrule
Word length (characters)  &  4.91 ± 0.04 & 4.72 ± 0.04 & *** \\
Word frequency (Wordfreq)  &  11.26 ± 0.08 & 10.98 ± 0.08 & *** \\
Word surprisal (Pythia 70M)  &  5.01 ± 0.06 & 4.77 ± 0.06 & *** \\
\bottomrule
\end{tabular}%
\label{table:text-stats}
\end{table}

\subsubsection{Participants} OneStopL1\&L2 has 638 adult participants, 360 English L1 participants in OneStop and 278 English L2 participants in OneStopL2. 334 of the participants read texts for comprehension. The remaining 304 participants read in an information-seeking regime described below. 
The mean participant age of L1 participants is 22.8, and their mean English age of acquisition (AoA) is 0.4. 
The mean participant age of L2 participants is 29.6 and their mean English AoA is 9.6. The L2 participants have the following 10 L1 backgrounds, with the number of participants per L1 in parentheses: Arabic (25), Chinese (34), French (23), Hebrew (32), Japanese (20), Korean (31), Portuguese (35), Russian (34), Spanish (35), and Vietnamese (9).

\subsubsection{Experimental setup} Each participant is assigned to one of three 10-article batches (54 paragraphs) in one of two between-subject reading regime conditions, ordinary reading for comprehension or information seeking. The texts are presented paragraph by paragraph. After each paragraph, the participant has to answer one of the three multiple-choice reading comprehension questions for the paragraph on a new screen, without the ability to return to the paragraph.
In the information-seeking regime, the question (but not the answers) is also presented before reading the paragraph.
Each paragraph in a given article is shown to the participant in either the original or simplified version, selected at random. The data further includes two articles in a repeated reading regime, which we do not use in our analyses.

The data is counterbalanced such that each participant reads 27 original and 27 simplified paragraphs overall and approximately the same number of original and simplified paragraphs within each article. 
On average, each paragraph is read by 
212 participants, 
106 in the original difficulty level and 
106 in the simplified level. 
Overall, the eye tracking data contains 3,733,817
 word tokens over which eye movement data was collected, 2,063,741 for the original texts and 1,670,076 for their simplified versions.

\subsubsection{Simplification effects in the data} An important prerequisite for the applicability of reading facilitation-centered evaluations is the presence of simplification effects on reading measures in the dataset. This question was investigated for L1 readers in OneStop by \citet{gruteke-klein-etal-2025-simplification}, who found a robust effect of simplification on reading times of 12\,ms in per word Total Fixation Duration, and significant reading speed increases in 42\% of the participants. The effects are even more pronounced in OneStopL2, where 69\% of the participants show significant reading speed increases and an overall effect of 32\,ms in per word Total Fixation Duration. These results suggest that both the L1 and the  L2 parts of the data are suitable for the development of reading facilitation-centered evaluations.

\subsection{Online Reading Ease Measures} 
Eye movements in reading are saccadic. Their scanpath trajectory consists of times in which the gaze is stable at a given location, called \emph{fixations}, and very rapid transitions between fixations, called \emph{saccades} \citep{rayner1998,schotter2025beginner}. While many words receive a single fixation, some receive multiple fixations, and some are skipped. Most saccades go forward in the text, but some are
regressive saccades that go backward.

We use three primary online measures from the psycholinguistic literature: 
\begin{enumerate}
    \item \textbf{Total Fixation Duration (TF)} TF is the sum of all the fixation durations on a word. We analyze average TF, taking into consideration only words that were not skipped. 
    \item \textbf{Skip Rate (SR)} The fraction of words that were skipped (i.e., were not fixated).
    \item \textbf{Regression Rate (RR)} The number of backward saccades per word. 
\end{enumerate}

All three measures capture processing difficulty. Longer reading times and less skipping are associated with increased processing difficulty during reading \citep{rayner1998,kliegl2004length,smith2013effect}, and with lower language proficiency \citep{cop2015eye,berzak2023traces} and lower reading proficiency \citep{mcconkie1991children,reichle2013using,tiffin2015word}.
Increased regression rates were similarly shown to be a marker of processing difficulty and sentence reanalysis \citep{clifton2007eye,staub2011word,bicknell2011readers,timkey2025preprint}. 
Correspondingly, more readable texts should have shorter Total Fixation times, more word skipping, and fewer regressions.

\subsection{Readability Scoring Methods} 

We evaluate 20 widely used traditional and modern readability scoring methods, commercial systems and LLM prompting. We compare them against key processing difficulty measures from the psycholinguistic
literature.

\subsubsection{Traditional formulas} 

We include 6 prominent and widely adopted measures from the readability literature: Flesch Reading Ease Score \citep{flesch1948new}, Dale-Chall Score \citep{dale1948formula}, Gunning Fog Index \citep{gunning1952technique}, Automated Readability Index (ARI) \citep{ari-smith1967}, Coleman-Liau Index \citep{coleman1975computer} and the Flesch Kincaid Grade Level \citep{flesch-kincaid1975}. These are linear regression formulas that use a small set of word property features, with coefficients fitted using reading comprehension data from English L1 speakers. Two commonly used features in these formulas are word length, a heuristic for measuring word complexity, and sentence length, a heuristic measure of grammatical complexity.  
Appendix~\ref{sec:app-readability_measures} Table~\ref{tab:readability_measures} presents the formulas and their interpretations. While these formulas were originally developed for passages, they are all applicable to single sentences. To compute these measures, we used the formula implementations in the \texttt{textstat} library version 0.7.4.

\subsubsection{Modern NLP methods for readability scoring}

Over the past few decades, the rise of modern NLP has led to the introduction of readability scoring methods that take advantage of automated linguistic analysis of texts and supervised machine learning.

\vspace{1em} 
\textbf{Supervised methods}
\vspace{0.5em} 

We selected several supervised NLP readability prediction methods that fulfill two key criteria. The first is their ability to provide a large range of readability scores, which supports a fine-grained characterization of the readability level of the text. We therefore do not examine methods whose output is restricted to a small set of categories (e.g., binary classifiers into two text difficulty levels). The second is that the method's code has to be publicly available. With these criteria in mind, we examine the following systems.

\begin{itemize}

\item{\textit{Coh-Metrix L2 Reading Index (CML2RI)}} \citep{cml2ri-crossley2008} A linear regression formula with three features, word frequency, sentence syntactic similarity and word overlap between adjacent sentences. The regression coefficients of CML2RI were fitted using L2 reading comprehension (cloze) scores. 

\item{\textit{Crowdsourced Algorithm of Reading Comprehension (CAREC)}} \citep{carec-cares-rossley2019} A linear regression formula from 13 linguistic features to a continuous readability score. The model is trained on readability scores derived using a Bradley-Terry model from L1 speakers' pairwise judgments of text difficulty (``Which text is easier to understand?''), for pairs of different texts.
    
\item{\textit{Crowdsourced Algorithm of Reading Speed (CARES)}}  \citep{carec-cares-rossley2019} A linear regression formula from 9 linguistic features to a continuous readability score. The model is trained on readability scores derived using a Bradley-Terry model from L1 speakers' pairwise judgments of reading speed (``Which text did you read more quickly?''), for pairs of different texts.

\item{\textit{Sentence-BERT (SBERT)}} \citep{crossley2023using} A transformer-based readability assessment model which uses Sentence-BERT embeddings \cite{reimers2019sentence}. The model produces continuous readability scores. It is trained on the CommonLit Ease of Readability (CLEAR) Corpus \cite{crossley2021commonlit,crossley2023large}, which includes excerpts normed to a school grade range of 3rd to 12th grade.

\end{itemize}
We use the implementation of the above methods in the Automatic Readability Tool for English (ARTE) \citep{choi2022advances}.
Similarly to traditional formulas, modern scoring methods were often developed based on data from passages, but can be applied to single sentences. 

\vspace{1em} 

\textbf{Unsupervised methods}
\vspace{0.5em} 

We additionally evaluate two unsupervised scores commonly used in NLP. 

\begin{itemize}    
    \item \textit{Perplexity (PPL)} A standard sequence probability score under a language model, defined for a word sequence $S$ as $PPL(S) = p(S)^{-\frac{1}{|S|}}$. Word probabilities are computed using the Pythia 70M language model \citep{biderman2023pythia}.
    
    \item \textit{Syntactic log-odds ratio (SLOR)} A measure that was originally proposed for modeling sentence acceptability \citep{pauls-klein-2012-large}, and was shown to correlate well with human sentence fluency ratings \citep{von2018sentence}. The measure is defined for sentence $S$ as: $\mathrm{SLOR}(S) = \frac{1}{|S|} \left( \ln p(S) - \ln p_u(S) \right)$, where $p(S)$ is the probability of the sentence under a language model, and $p_u(S)$ is the unigram probability of the sentence. 
    Sentence probabilities are computed using Pythia 70M \citep{biderman2023pythia}. Unigram probabilities are estimated using word frequencies from  Wordfreq \citep{robyn_speer_2018}. To compute the SLOR score of a passage, we average the SLOR scores of the sentences that comprise it.

\end{itemize}

\subsubsection{Commercial readability systems} 
Readability scoring of texts is widely used in real-world settings. It is therefore important to further consider readability scoring systems that are used predominantly in such contexts. Here, we examine two prominent commercial systems for readability scoring, which are used in educational settings.

\begin{itemize}

\item{\textit{Lexile Text Analyzer}} A commercial readability assessment system developed by MetaMetrics (\href{https://hub.lexile.com/text-analyzer/}{https://hub.lexile.com/text-analyzer/}), which is commonly used in the US K-12 educational system (\href{https://metametricsinc.com/products/state-eogeoc-assessments/}{https://metametricsinc.com/products/state-eogeoc-assessments/}). In the spirit of traditional readability formulas, Lexile readability scoring is based on regression from word frequency (``semantic component'') and sentence length (``syntactic component'') to reading comprehension outcomes \citep{lexile2022}. The score range is 0 to 2000. 
Lexile scores were obtained using the Lexile API, accessed on April 27, 2025.

\item{\textit{TextEvaluator}} A commercial readability scoring system by the Educational Testing Service (ETS) (\href{https://textevaluator.ets.org/TextEvaluator/}{https://textevaluator.ets.org/TextEvaluator/}). The system uses a variety of textual features extracted with NLP tools, regressed against human annotations of text readability level \citep{sheehan2014textevaluator}.    
TextEvaluator scores range from 100 to 2000. The scores were obtained by manually querying the TextEvaluator web interface, accessed on April 27, 2025.

\end{itemize}

\subsubsection{Large Language Models Prompting} 
\label{sec:methods-llms}

LLMs have revolutionized NLP, and among their many uses, they are increasingly used in readability research and applications \cite[][among others]{trott-riviere-2024-measuring, farajidizaji-etal-2024-possible, creutz-2024-correcting}. 
In the current study, we evaluate 6 current-frontier LLMs through their APIs. \begin{itemize}
    \item \textit{Llama 3.3 70B Versatile} (Meta, 2024).
    \item \textit{GPT-4o}: \texttt{gpt-4o-2024-08-06} (OpenAI, 2024).
    \item \textit{GPT-5}: \texttt{gpt-5-2025-08-07} (OpenAI, 2025).
    \item \textit{Gemini 2.0 Flash}: \texttt{gemini-2.0-flash-001} (Google, 2025).
    \item \textit{Gemini 2.5 Pro}: \texttt{gemini-2.5-pro} (Google, 2025).
    \item \textit{Claude Sonnet 4}: \texttt{claude-sonnet-4-20250514} (Anthropic, 2025).
\end{itemize}

The Llama 3.3 model is open-weights, while the remaining 5 models are closed-source. We evaluate these models using two different annotation frameworks provided to the models in the prompt: 
\begin{enumerate} 
    \item \textit{Grade-level}: predict a school grade level (1--12). School grades are commonly used for readability level annotation \citep{vajjala2022trends}.
    \item \textit{Score}: assign a readability score from 1 (easy) to 100 (hard), following \citet{trott-riviere-2024-measuring}.
    \end{enumerate}
Following \citet{trott-riviere-2024-measuring}, we also experiment with a second variant for each of the prompts above, which includes additional guidance for scoring the text, to consider sentence structure, discourse structure, vocabulary, and clarity. 
For the main analysis, we use the \textit{Grade-level} prompt with additional guidance. The other prompts yield similar results.
The prompts and the results for all four prompt variants are provided in Appendix~\ref{sec:prompt_variants}.

\subsection{Psycholinguistic Measures}
\label{sec:psycholing-measures}

\citet{howcroft-demberg2017} introduced the idea of using theoretically motivated measures from the psycholinguistic literature, which have been tied to memory and processing load for predicting text readability. They used 4 such measures, idea density, integration cost, embedding depth and surprisal, for the classification of sentence difficulty level for pairs of original and simplified sentences. 
Here, we examine the extent to which these and additional psycholinguistic measures are predictive of reading facilitation as a result of simplification. These evaluations provide a key benchmark for the evaluations of the readability scoring methods presented above.

\vspace{1em} 
\textbf{Memory related measures}
\vspace{-1em}
\begin{itemize}
\item{\textit{Idea Density}} The ratio of ideas or propositions to words \citep{kintsch1972notes}. 
This ratio is expected to be lower in more readable texts. To compute idea density, we use the Computerized Propositional Idea Density Rater (CPIDR) 3.2 \citep{brown2008automatic}, which approximates idea density as the ratio between the number of part-of-speech tags that typically introduce new information and the total number of words. We computed this measure using the \texttt{cpidr} function implemented in the \texttt{pycpidr} library version 0.3.0.
    
\item{\textit{Integration Cost}} This measure is rooted in the dependency locality theory (DLT) \citep{gibson1998linguistic,gibson2000dependency}, which posits that greater distance between syntactically related words increases the difficulty of linking them in memory. Integration cost is defined via the distance between syntactic heads and dependents, measured by the number of intervening discourse referents. 
More readable texts should have shorter dependencies and correspondingly, a lower average integration cost. We use \texttt{icy-parses} to compute for each sentence the average integration cost per word.  
To obtain the integration cost for a passage, we calculate the average integration cost of its sentences.
    
\item{\textit{Embedding Depth}} Syntactic embedding depth is another measure that reflects expected processing difficulty due to memory load and corresponds to the number of open dependencies a human parser needs to maintain at a given point in time. 
Following \citet{howcroft-demberg2017}, we compute embedding depth at a given word using the incremental left-corner parser from \citet{van2012connectionist}. 
We compute the average embedding depth for a passage as the average embedding depth of its sentences.
\end{itemize}

\textbf{Information-theoretic measures}
\vspace{-1em}
\begin{itemize}
\item{\textit{Surprisal}} Surprisal theory \citep{hale2001,levy2008} relates processing difficulty to the predictability of the word in its context. Surprisal is defined as $-\log_2(p(w|context))$, where \textit{context} denotes the words preceding the current word \textit{w} in the word sequence $S$. The average surprisal for a word sequence is: 
$ \text{surp\_avg(S)} = 
    \frac{1}{|S|} \sum_{w \in S} -\log_2(p(w|context))
$. 
In the main analysis, we use the Pythia 70M language model \citep{biderman2023pythia}, whose surprisal estimates were shown to correlate well with reading times \citep{gruteke-klein-etal-2024-effect}. 
Following standard practice \citep[e.g.,][]{oh2023does,shain2024large}, we compute the surprisal of a word by summing the surprisals of its subword tokens.

\end{itemize}

The following are additional information-theoretic measures not included in \citet{howcroft-demberg2017}. 

\begin{itemize}    
\item{\textit{Pseudo Log Likelihood (PLL)}} A method for scoring words under bidirectional Masked Language Models (MLMs) \citep{wang2019bert}. While less commonly used in reading research, we include this method because it is related to surprisal and is conceptually similar to bidirectional cloze tasks, which were originally suggested for readability scoring \citep{taylor1953cloze}. We use PLL-word-l2r \citep{kauf2023better}, with \texttt{minicons} \href{https://github.com/kanishkamisra/minicons}{https://github.com/kanishkamisra/minicons}, with the best performing model from \citet{kauf2023better}, \texttt{roberta-large}.

\item \textit{Uniform Information Density (UID)} According to the uniform information density hypothesis \citep{jaeger2006speakers}, language comprehenders prefer information to be more uniformly distributed across words. Here, we use the \textit{Global Variance (Language)} variant of UID from \citet{meister2021revisiting}, which is defined as $\mathrm{UID}^{-1}(S) = \frac{1}{|S|} \sum_{w \in \text{S}}  (surprisal(w) - \mu)^2$,  where $w$ is a word in the sequence $S$, and $\mu$  is an approximation of the mean word surprisal in the language, based on the WikiText-103 corpus. Surprisal is estimated using Pythia 70M \citep{biderman2023pythia}.
For passages, we compute the average UID across sentences. 

\item{\textit{Entropy}} of the next word given a context is another information-theoretic measure that was linked to processing difficulty. Shannon word entropy is defined as $-\sum_{w \in V}{p(w|context)\log_2(p(w|context))}$, where $V$ is the vocabulary. We use mean per-word entropy. We include this measure as it has been shown to predict reading times above and beyond surprisal \citep{cevoli2022prediction,pimentel2023effect}. We use the Pythia 70M language model \citep{biderman2023pythia} to compute word probabilities. Similarly to surprisal, we estimate the entropy of a word by summing the entropy values of its subword tokens. The resulting quantity is an upper bound of word entropy, whose exact computation is intractable \cite{bolliger-etal-2024-alignment}.

\end{itemize}

In addition to the above measures, similarly to \citet{howcroft-demberg2017}, we include average word length and average word frequency. Together with predictability (measured using surprisal), these measures form the ``big three'' word property predictors of reading times \citep{clifton2016eye}.

\begin{itemize}

\item{\textit{Word Length}} is a strong predictor of reading times and skipping behavior in reading \cite[e.g.,][]{brysbaert1998word,joseph2009word,kliegl2004length,rayner2011eye}. It is also a common feature used in traditional readability measures (see Appendix~\ref{sec:app-readability_measures} Table~\ref{tab:readability_measures}). We measure word length in characters, excluding punctuation.

\item{\textit{Word Frequency}} is similarly a robust predictor of reading times, which has been tied to lexical processing during reading  \cite[e.g.,][]{rayner2004effects,kliegl2004length,shain2024word, berzak2023traces}. We use frequency counts from Wordfreq \citep{robyn_speer_2018}, coded as unigram surprisal $-\log_2(p(w))$. 
\end{itemize}

We use the \texttt{psycholing-metrics} library version 1.1.7 \citep{yoav_meiri_2026_20376826} for calculating per-word surprisal, entropy, length, and frequency. Additional details on the libraries used for computing readability scores and psycholinguistic measures are provided in Appendix~\ref{sec:implementation}. Pairwise Pearson correlations between all the readability measures based on \mbox{OneStopL1\&L2} sentences and passages are presented in Appendix~\ref{sec:readability_measures_correlations} Figure~\ref{fig:all_readability_measures_correlations}.

\subsection{Evaluation Framework} Our evaluation criterion for readability scoring methods is their ability to account for the \emph{reading facilitation} that readers experience as a result of \emph{text simplification}. Importantly, this criterion provides experimental control for the content of the texts. We capture this facilitation effect as the change in reading ease measures between the original and the simplified version of each text:
\begin{align}
\Delta \text{ReadingEase}_{E,T} 
&= \text{ReadingEase}_{E,T_{\text{original}}} - \text{ReadingEase}_{E,T_{\text{simplified}}} \nonumber
\end{align}
where $E$ is an eye movements measure  $\in \{ \text{TF}, \text{SR}, \text{RR}\}$, $T$ is a textual item,   $\text{ReadingEase}_{E,T_{\text{original}}}$ is the average reading ease according to measure $E$ across participants reading the original version of the text,
and $\text{ReadingEase}_{E,T_{\text{simplified}}}$ is the average of the same measure for the participants reading the simplified version of the same text.  In our main analysis $\text{ReadingEase}_{E,T_{\text{original}}}$ and $\text{ReadingEase}_{E,T_{\text{simplified}}}$ are each based on 106 participants on average. $T$ can be a sentence or a passage.

We further define a corresponding difference in readability scores for the same text, according to a readability scoring method $M$:
\[\Delta \text{Score}_{M,T} =  \text{Score}_{M,T_{\text{original}}} -  \text{Score}_{M,T_{\text{simplified}}}
\]
To evaluate the quality of a readability scoring method $M$, we measure the predictivity of $\Delta \text{Score}_{M,T}$ for $\Delta \text{ReadingEase}_{E,T}$ for the same texts using Pearson correlation $r$:
\[
\text{Eval}_{\text{M}} = \text{Pearson}\_{\text{corr}}(\Delta \text{Score}_{M,T}, \Delta \text{ReadingEase}_{E,T})
\]

\section{Results}

\subsection{Main Analysis}

\begin{figure*}[ht!]
    \centering
    \includegraphics[width=1\textwidth]{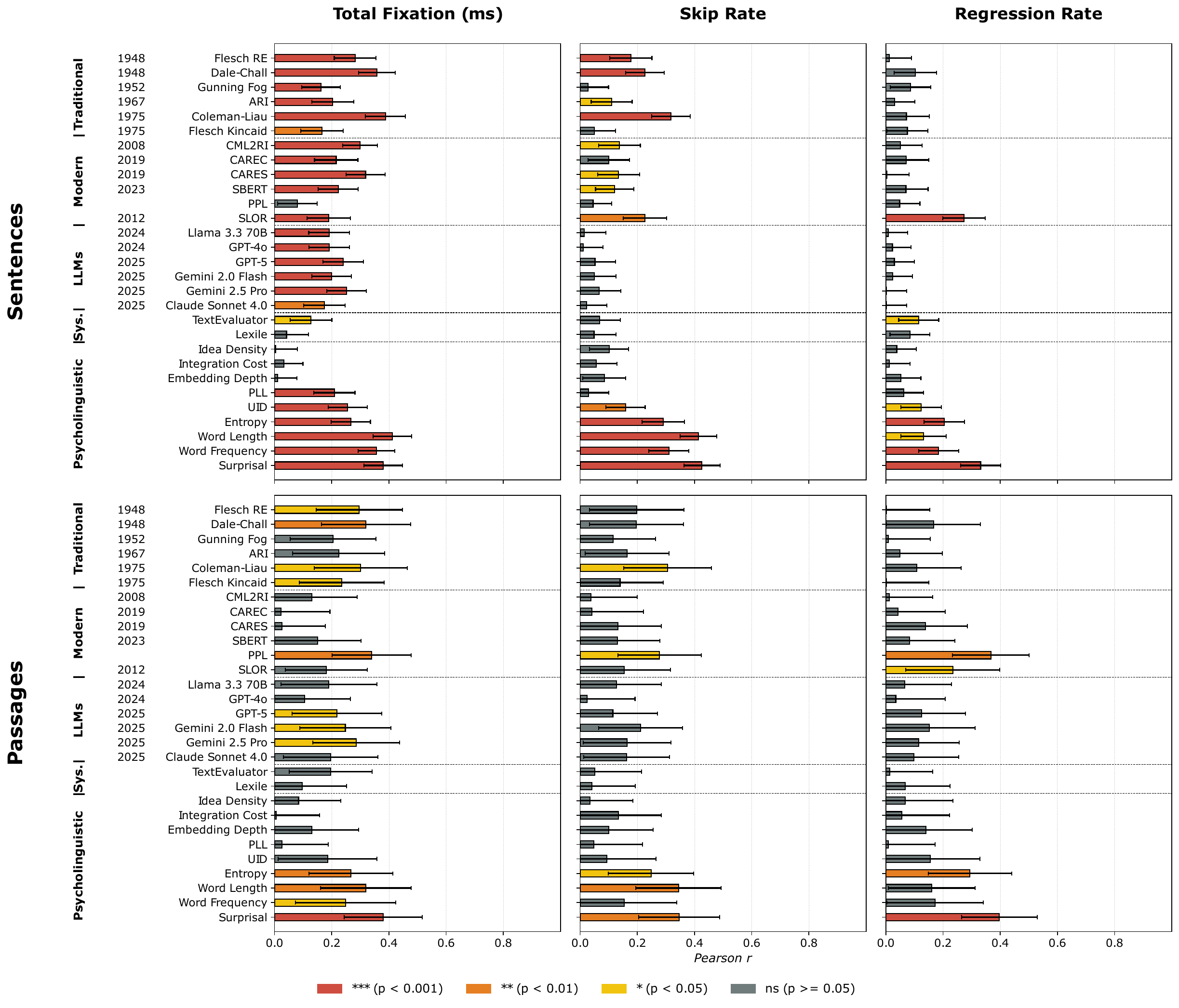}
    \caption{
\textbf{Reading ease predictivity of readability scoring methods and psycholinguistic measures.} 
Years denote the publication or model release year of readability scoring methods. The evaluations use three eye movement measures of reading ease: Total Fixation (the sum of all fixation durations on fixated words), Skip Rate (the fraction of skipped words), and Regression Rate (the number of backward saccades per word). 
Each bar depicts the Pearson correlation $r$ coefficient of $\Delta \text{Score}_{M,T}$ with $\Delta \text{ReadingEase}_{E,T}$, where $\Delta \text{Score}_{M,T}$ is
the readability score difference between pairs of original and simplified texts $T$ according to method $M$, 
and $\Delta \text{ReadingEase}_{E,T}$ is the average reading ease difference according to measure $E$ on the same pairs of texts across participants. 
Error bars represent 95\% confidence intervals using a bootstrap over textual units (200 resamples with replacement). 
Colors represent the statistical significance level of the correlation. }
\label{fig:RTxSenPar_pearson_corr_FirstReading_boot}
\end{figure*}

The results of our main analysis using the entire OneStopL1\&L2 corpus are presented in \Cref{fig:RTxSenPar_pearson_corr_FirstReading_boot}. Depicted are the Pearson correlations for readability scoring methods and psycholinguistic measures with $\text{Total Fixation}$, $\text{Skip Rate}$ and $\text{Regression Rate}$, at the sentence and passage levels. Results of the corresponding pairwise statistical comparisons between the correlation coefficients of all the methods and psycholinguistic measures are presented in 
Appendix~\ref{sec:main_analysis_pairwise_stat_comparisons} Figure~\ref{fig:RT_perm_test_grid_RTxSenPar_boot_FirstReading}. Additional details on statistical testing and comparisons of correlation coefficients are provided in Appendix~\ref{sec:pairwise_statistical_comparisons}. 

The traditional, modern supervised NLP, and LLM-based readability methods tend to have low and, in many cases, non-significant correlations across both passages and sentences, and different reading ease measures. Notably, despite the increasing sophistication of the underlying models, we do not observe a pattern of improvement in readability scoring method performance over time. Especially low predictivity is obtained for the two commercial systems, TextEvaluator and Lexile.

Among the psycholinguistic measures, the memory-related measures, idea density, integration cost, and embedding depth, as well as PLL, do not yield significant correlations in most cases. Differently from these predictors, the information-theoretic measures entropy and (for sentences) UID, as well as the big three, length, frequency, and surprisal, are on par with or better than the highest performing readability scoring methods in most evaluations. 

The main competitors of the big three and entropy among the traditional formulas are Dale-Chall and Coleman-Liau. This is likely related to the strong correlation of these two formulas with word length 
(see Appendix~\ref{sec:readability_measures_correlations} Figure~\ref{fig:all_readability_measures_correlations}). In the unsupervised NLP category, PPL performs well on passages and SLOR on sentences. This outcome is in line with the direct relation of these measures to frequency and surprisal; PPL is closely related to surprisal, and SLOR is a combination of surprisal and frequency.
The advantage of the big three and entropy is especially apparent in the Regression Rate evaluation, where the correlations of nearly all other methods except for PPL and SLOR are not significant. The best predictor overall tends to be surprisal. 

These results are consequential. They suggest that although widely adopted in research and real-world settings, prominent readability scoring methods are very limited in their ability to account for reading ease. These methods are outperformed by simple word properties commonly used for reading time prediction in psycholinguistics.

\subsection{Analyses of Generality}

How general and robust are the results of the main analysis? Below, we take several steps in addressing this question by providing analyses that test the extent to which they depend on the reader population (native versus non-native English speakers), reading regime (reading for comprehension versus information seeking), as well as several key methodological choices in the experimental setup. 

\subsubsection{Reader groups, reading regimes and textual units} Prior work has shown differences in reading patterns across L1 and L2 \cite[e.g.,][]{whitford2012second,cop2015eye,berzak2023traces}. Different reading patterns were also observed across different reading regimes, in particular between reading for comprehension and information seeking \cite[e.g.,][]{kaakinen2015influence,hahn2023modeling,shubi2023eye}. 
Here, we examine whether such differences are consequential for our evaluation.

In Appendix~\ref{sec:L1_L2_separate} Figure~\ref{fig:SM_RT_main_RTxSenPar_pearson_corr_boot_L1_next_to_L2_FirstReading} we present the evaluations separately for the L1 and L2 participants, where we observe largely similar results across the two groups. Similarly, consistent results are obtained in Appendix~\ref{sec:information_seeking} \Cref{fig:SM_RT_main_RTxSenPar_pearson_corr_boot_Hunting0_next_to_Gathering0} when separately examining ordinary reading for comprehension and information seeking.
These evaluations suggest a key strength of our proposed reading ease focused evaluation framework: it is largely invariant both across different reader groups and across different reading regimes. Finally, we note that qualitatively similar result patterns are obtained for sentence-level and passage-level predictivity across the different analyses presented in the Appendix.

\subsubsection{Additional reading ease measures} 
The main analysis uses Total Fixation, Skip Rate and Regression Rate as the primary reading measures of interest.
In Appendix~\ref{sec:additional_RT_measures}, we 
present the same analysis for eight additional eye tracking measures from the psycholinguistic literature, as well as for reading speed.

\vspace{1em} 
Single fixation measures and number of fixations (Appendix~\ref{sec:additional_RT_measures} Figure~\ref{fig:SM_RT_1_RTxSenPar_pearson_corr_boot_FirstReading}):
\begin{itemize}
    \item \textbf{First Fixation (FF)} The mean duration of the first fixation on a word, considering only words that were not skipped.
    \item \textbf{Fixation Duration (FD)} The mean duration of a fixation on a word.
    \item \textbf{Number of Fixations (NF)} The mean number of fixations per word.
\end{itemize}

First pass measures (Appendix~\ref{sec:additional_RT_measures} Figure~\ref{fig:SM_RT_2_RTxSenPar_pearson_corr_boot_FirstReading}):
\begin{itemize}
    \item \textbf{First Pass Gaze Duration (fpGD)} The mean time from first entering a word to first leaving it, during first pass reading.
    \item \textbf{First Pass Skip Rate (fpSR)} The fraction of words that were not fixated during first pass reading.
    \item \textbf{First Pass Regression Rate (fpRR)} The number of saccades per word that go backward, during first pass reading.
\end{itemize}

Later measures and reading speed (Appendix~\ref{sec:additional_RT_measures} Figure~\ref{fig:SM_RT_3_RTxSenPar_pearson_corr_boot_FirstReading}):
\begin{itemize}
    \item \textbf{Gaze Duration (GD)} The mean time from first entering a word to first leaving it. 
    \item \textbf{Higher Pass Fixation Duration (hpFD)} The sum of all fixation durations on a word during second and higher pass readings, averaged across words. 
    \item \textbf{Reading Speed (RS)} The number of words read per second. Note that, differently from the measures above, RS is an offline measure that can be obtained without eye tracking. Assuming the text is known, it requires only the total reading time of the text.    
\end{itemize}

The main results largely hold for these additional measures. 
\subsubsection{Correlation measure} 
In our main analysis, the correlations between text readability measures and reading ease measures are quantified using the Pearson $r$ coefficient, under an assumption of linearity. However, this assumption may not hold in all cases, which could affect the interpretation of the results.
In Appendix~\ref{sec:spearman} Figure~\ref{fig:SM_RT_main_RTxSenPar_pearson_next_to_spearman_corr_boot_FirstReading} we present the main analysis using  Spearman $\rho$, which relaxes this assumption. We find that the results are qualitatively similar to those obtained with Pearson $r$. 

\subsubsection{Annotation prompts for LLMs} 

LLM performance can depend on the annotation framework, as well as on the wording used to formulate the task in the prompt \citep[][among others]{loya2023exploring,razavi2025benchmarking}.
As described in Section~\ref{sec:methods-llms}, we experiment with two different readability level annotation schemes, each with two different instruction formulations provided to the LLMs via the prompt. In Appendix~\ref{sec:prompt_variants}
we present the prompts and their corresponding performance on our evaluation. The results are largely consistent across the different prompt variants.

\subsection{Comparison to Reading Ease without Control for Text Content} 
\label{sec:comparison-with-direct-pred}

A key aspect of our proposed framework is the utilization of a parallel corpus of original and human-simplified texts, which allows for content-controlled evaluations focusing on reading facilitation as a result of simplification. How does this approach compare to the direct prediction of reading times commonly performed in psycholinguistics, and also used in the context of readability \citep{nahatame2021text} (see discussion in Section~\ref{sec:discussion})? 
In the direct prediction approach, the quality of a readability scoring method $M$ is measured using the predictivity of $\text{Score}_{M,T}$ for $\text{ReadingEase}_{E,T}$ with different texts.
\[
\text{Eval}_{\text{M}} = \text{Pearson}\_{\text{corr}}(\text{Score}_{M,T}, \text{ReadingEase}_{E,T})
\]

To address this question, in Appendix~\ref{sec:all_levels_results} Figure~\ref{fig:SM_all_levels_pearson_corr_RE_next_to_delta_RE_boot_FirstReading} we present a comparison between an evaluation of our main analysis and the direct prediction of reading ease measures for all the texts in the OneStopL1\&L2 data. We observe that the lack of control for content does affect the evaluation outcomes: for most readability scoring methods, and especially the traditional formulas, the content-controlled evaluations yield lower correlations compared to the corresponding evaluations without such control. This suggests that these methods are substantially affected by the text topic, i.e., by factors that go beyond writing style. A similar trend is observed for word length and frequency. Surprisal, UID, and entropy, on the other hand, tend to obtain stable results across both evaluation methods. 

\subsection{Surprisal as a Readability Measure}
\label{sec:surp-different-lms}

\begin{figure*}[ht!]
    \centering
    \includegraphics[width=1\textwidth]{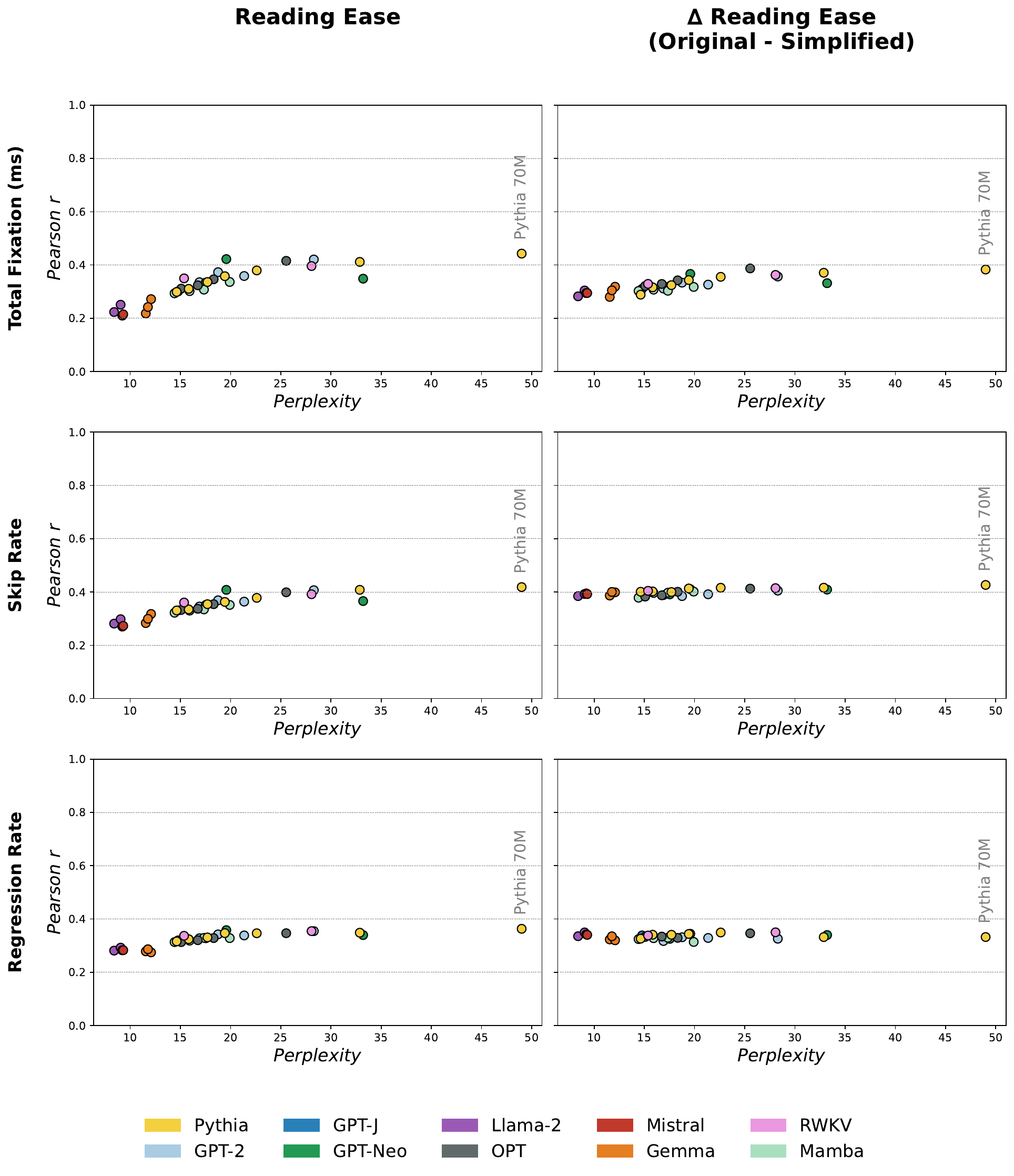}
    \caption{
\textbf{Surprisal: robustness of reading ease predictivity to the choice of language model.} 
Analysis at the \textbf{sentence} level. 
Left: direct prediction of reading ease without text content control,
$\text{Eval}_{\text{Surprisal,LM}_i} = \text{Pearson}\_{\text{corr}}\big( \text{Surprisal}_{\text{LM}_i,T}, \text{ReadingEase}_{E,T} \big)$ where $\text{Surprisal}$ are mean word surprisal values per sentence according to language model $\text{LM}_i$. Right: our proposed evaluation with text content control, 
 $\text{Eval}_{\text{Surprisal,LM}_i} = \text{Pearson}\_{\text{corr}}\big(\Delta \text{Surprisal}_{\text{LM}_i,T}, \Delta \text{ReadingEase}_{E,T} \big)$. 
Colors represent the model family. Model sizes range from 70 million to 13 billion parameters.
The main analysis uses Pythia 70M.}
\label{fig:RTxLevel_pearson_corr_by_perplexity_sentence_FirstReading}
\end{figure*}

Surprisal emerged as the measure that tends to account best for reading ease under our evaluation method. Our final analysis provides a detailed examination of this measure, which combines the two analysis types above to examine the stability of this measure within and across evaluation methods, as a function of the choice of language model from which the surprisal values were obtained. In Section~\ref{sec:comparison-with-direct-pred} above, we have already observed that for Pythia 70M, the predictivity of surprisal is similar across the two evaluation methods. Our focus here will be on the sensitivity of the evaluations to the model choice. This focus is motivated by converging evidence in the psycholinguistic literature, which suggests that in recent language models, lower perplexity models exhibit reduced predictive power for reading measures~\cite[among others]{oh2023does,shain2024large,gruteke-klein-etal-2024-effect}. 

To enable this investigation, in addition to Pythia 70M, we extracted surprisal values for OneStopL1\&L2 from 31 additional public language models from the GPT-2 \citep{radford_language_2019}, GPT-J \citep{gpt-j}, GPT-Neo \citep{gpt-neo}, Pythia \citep{biderman2023pythia}, OPT \citep{zhang2022opt}, 
Mistral \citep{jiang2023mistral7b}, Gemma \citep{team2024gemma}, 
Llama-2 \citep{touvron2023llama}, RWKV \citep{peng2023rwkv}, and Mamba \citep{gu2024mamba} families, ranging from  70 million to 13 billion parameters. The complete list of models is provided in Appendix~\ref{sec:robust_LLM} Table~\ref{tab:models_table}. 

In Figure~\ref{fig:RTxLevel_pearson_corr_by_perplexity_sentence_FirstReading}, on the left, we compare the predictivity of surprisal for sentence reading ease without controlling for content (i.e., using the direct prediction approach) across different language models, as a function of model perplexity.
We observe that reading ease predictivity indeed increases substantially as the models' perplexity increases across all three reading ease measures: ($\beta=0.0062$, $p<10^{-7}$) for $\text{Total Fixation}$, ($\beta=0.0040$, $p<10^{-8}$) for $\text{Skip Rate}$, and ($\beta=0.0024$, $p<10^{-6}$) for $\text{Regression Rate}$. Similar results are obtained at the passage level in Appendix~\ref{sec:robust_LLM} Figure~\ref{fig:RTxLevel_pearson_corr_by_perplexity_paragraph_FirstReading}.

In Figure~\ref{fig:RTxLevel_pearson_corr_by_perplexity_sentence_FirstReading}, on the right, we perform the same analysis using our proposed content-controlled evaluation. Differently from direct prediction, here, we obtain only moderate predictivity increases as a function of model perplexity for $\text{Total Fixation}$ ($\beta=0.0027$, $p<10^{-7}$) and $\text{Skip Rate}$ ($\beta=0.001$, $p<10^{-5}$), and no increase for $\text{Regression Rate}$ ($\beta=0.00006$, $p>0.05$).  
Importantly, the increases are significantly smaller than in the direct prediction setting ($p<0.001$ in all cases, assessed using the $\text{model\_perplexity:evaluation\_method}$ interaction term in the model 
$ Pearson\_corr \sim model\_perplexity * evaluation\_method
$). 
In Appendix~\ref{sec:robust_LLM} Figure~\ref{fig:RTxLevel_pearson_corr_by_perplexity_paragraph_FirstReading}, we obtain similar results for passages.

Thus, the relative stability of surprisal evaluations across different LMs is an additional advantageous byproduct of the content-controlled evaluation setting. More broadly, although surprisal is not often used as a readability measure, the combination of its strong results and stability across different models underlines its value as a simple and general measure of readability.

\section{Discussion}
\label{sec:discussion}

Departing from a century-old tradition in ARA, which predominantly relies on reading comprehension outcomes and readability level annotations, we focus on the construct of reading ease, and introduce reading facilitation as a result of simplification as a cognitive benchmark for the evaluation of readability scoring methods. We use this framework to evaluate a wide range of prominent readability scoring approaches, which we further benchmark against key psycholinguistic measures that have been previously theoretically or empirically linked to online processing difficulty. Based on this analysis, we find that widely used readability scoring methods are poor predictors of reading ease, outperformed by word length, frequency, and two information-theoretic predictors, entropy and surprisal. 

Our work adds to a relatively small body of work that used behavioral data in the context of readability assessment.
A notable example of early work in this area is \citet{Mcclusky1934AQA}, who used words read per second (WPS) as a ground truth measure for text difficulty. In another early study, \citet{clare1957} used an eye tracking device to obtain measures of WPS and words read per fixation, comparing these measures for easy and hard versions of the same passages. \citet{kintsch-vipod} present unpublished data from Ernst Rothkopf, who used eye movement measures (number of regressions, fixations and saccades) for estimating the readability of texts from different genres. 

More recently, \citet{vajjala-etal-2016-towards} and \citet{gruteke-klein-etal-2025-simplification} studied differences in eye movement measures between original and simplified versions of the same texts. Here, we leverage such differences to control for topical variation in the assessment of readability scoring methods.  \citet{nishikawa2013pilot} conducted a pilot study for predicting readability from textual features, where the gold readability label of a document is its overall reading time. In a similar vein, \citet{scozzaro2024reform} trained a regression model that predicts word reading times, used as measures of readability. Differently from these two studies, which focus on using reading times for the readability evaluation of specific textual corpora, our work aims to abstract away from specific texts and instead evaluate the general quality of existing readability scoring methods. 

To our knowledge, the only two studies to date to perform regressions between ARA scores and reading measures are \citet{nahatame2021text} and \citet{hollenstein2022patterns}. 
The main differences between these studies and our work are our focus on the use of reading ease as a gold standard for the evaluation of ARA methods, and our content-controlled evaluation setting which allows controlling for topical variation between texts. Further additions of our study include the evaluation of LLMs, commercial readability systems, and additional readability formulas and psycholinguistic measures, including surprisal, which was not studied in these works. Differently from \citet{nahatame2021text}, we do not find improved reading time correlations in modern measures compared to traditional formulas.

Our results have several important implications for readability research and applications. First, they provide empirical evidence for the drawbacks of existing readability methodologies in capturing the real-time experience of readers with texts, and call for caution in the adoption of existing methods in high-stakes settings. This empirical evidence reflects both conceptual and practical limitations of reading comprehension and human annotation approaches for quantifying readability, which we discuss below. 

While comprehension ease is indeed central to readability, its practical assessments have a number of limitations. First, comprehension scores depend not only on the difficulty of the text, but also on the difficulty of the reading comprehension questions. This issue is further exacerbated when comparing the readability levels of different texts using different questions. Further, with a limited number of questions, it is impossible to probe reading comprehension exhaustively, and highly challenging to do so reliably. Fundamentally, it is a measure that cannot fully account for readability: differences in reading and comprehension ease do not have to lead to differences in reading comprehension performance, and vice versa, differences in reading comprehension performance do not necessarily stem from differences in text readability. Indeed, several studies found that substantial reading ease differences in adult L1 speakers translate to negligible differences in reading comprehension performance \citep{vajjala2019understanding,starc2020,gruteke-klein-etal-2025-simplification}.

The alternative, human readability labeling approach includes absolute and comparative labeling of the readability level of different texts (e.g., graded readers and pairwise annotation of relative readability). Level annotations have been obtained both from experts, such as teachers, and from non-expert annotators, for example, via crowdsourcing. While such annotations have been widely adopted for training and evaluating readability scoring systems in NLP, they too have fundamental drawbacks \citep{vajjala2022trends}. First, labeling is operationalized via highly non-trivial annotation tasks. Inter-annotator agreement on these tasks is largely unknown, and was found to be low when measured \citep{brunato2018sentence}. Further, annotations are often limited to a small set of discrete, annotation-framework-specific levels.

Eye tracking enables a viable alternative to the reading comprehension and labeling approaches, which obviates many of their limitations. Perhaps its key advantage is that, differently from comprehension questions and human judgments, which are both \emph{offline} behavioral signals, it provides \emph{online} measures of how readers experience the text. Unlike many human annotation frameworks, it offers continuous scores according to multiple interpretable measures. While in this work we highlight reading ease, such measures are also intimately related to online language comprehension processes and, as such, provide a multifaceted measure for readability. 

An additional long-standing challenge in readability assessment is the topical variability across texts. According to \citet{vajjala2022trends}, this ``... leads us to question what the ARA models learn -- is it a notion of text complexity, or topical differences among texts?''. In this work, we introduce a new method for controlling for such variability by focusing on the readability change as a result of largely content-preserving simplification edits. While in the current study we use this method in the context of reading ease, similar methodologies can also be applied in the future to annotation and comprehension-based readability benchmarks.

Despite their many advantages, eye tracking based methodologies also have limitations. First, they are not the only source of cognitive information that is relevant for readability. For example, post-reading text recall \citep{martinez2022poor} may also be of major interest. More broadly, reading ease is best viewed as an additional, thus far underexplored tool, which complements existing approaches. Eye movements also pose practical challenges. Currently, the collection of high-quality eye tracking data for reading is costly and difficult to scale \cite{jakobi2026eye}. Furthermore, such data includes substantial unexplained variability, as well as individual differences in eye movement patterns across readers \citep[][among others]{Calvo01072001, ashby2005eye, kim2022developmental, gruteke-klein-etal-2025-simplification}. In this work, we tackle these challenges by computing summary statistics over multiple words and averaging these statistics over a large number of participants. This requires collecting data from many participants, which is not always feasible. Summary statistics also involve the loss of potentially valuable information from the full eye movement trajectories.

We further acknowledge that similarly to comprehension and annotation-based methods, eye movement measures may capture characteristics of the reader, the text and their interaction, which go beyond reading ease. These include interest and engagement \cite{miller2015using,kaakinen2018fluctuation,rettig2023relations}, topic familiarity \cite{kaakinen2003prior,copeland2013effect}, reading strategies \cite{hyona2002individual,wotschack2009eye}, response to semantic word properties like valence and arousal \cite{scott2012emotion,knickerbocker2015emotion}, concreteness \citep{magnabosco2024eye}, and others. Our study includes a large number of text topics, a randomization of participant assignments to texts and text difficulty levels, and content control across the two text difficulty levels. However, it is conceivable that despite these controls, systematic changes in one or more of these factors are present across the two text difficulty levels. We leave the investigation of this question, which may require additional data collection for the reading aspects of interest, to future work.

The results of this study speak to the relevance of psycholinguistically rooted measures for text readability scoring. While memory-centered predictors based on idea density, integration cost, and embedding depth do not perform well on our evaluations, information-theoretic measures and standard word properties used for reading time prediction emerge as the strongest predictors of reading ease. The differences among the psycholinguistic predictors are in line with the more ample evidence in the psycholinguistic literature on the direct effects of information-theoretic predictors on reading times compared to memory-based predictors.

The overall best performing predictor, surprisal, is closely related to the unsupervised NLP scores PPL and SLOR. Specifically, $PPL(S) = 2^{\text{surp\_avg} (S)}$. The stronger performance of surprisal compared to PPL is likely related to converging evidence with multiple reading corpora \cite{smith2013effect,shain2024large,buggy2026predictability}, including OneStop \cite{gruteke-klein-etal-2024-effect}, which suggests that the relation between reading times and word probabilities is logarithmic rather than linear. SLOR can be expressed as $\text{unigram\_surp\_avg} (S)-\text{surp\_avg} (S)$. This combination of surprisal and frequency turns out to be less robust than using surprisal alone.
Surprisal is also conceptually related to left-to-right cloze \citep{smith2011cloze}. Although the current primary use of left-to-right cloze is the estimation of subjective probabilities in psycholinguistics, its bidirectional version was originally proposed as a method for quantifying readability \citep{taylor1953cloze}, and bidirectional cloze tests were used for fitting the regression coefficients of several traditional readability formulas \citep{ari-smith1967,flesch-kincaid1975,coleman1975computer}. Interestingly, PLL, which is a corpus-based approximation for bidirectional cloze, does not perform well, even compared to these traditional readability formulas. In addition to its weaker psycholinguistic basis, the performance of PLL could further be hindered by the inherent challenge of approximating human predictions in fill-in-the-blank tasks using language models. We leave further investigation of this question to future work.

Practically, our results support a broader adoption of surprisal for readability assessment. 
Unlike cloze, surprisal is not limited by the need to collect large-scale data from human participants, and can be computed automatically using readily available language models. More broadly, the results motivate additional work on eye movement based evaluation of ARA in other languages, populations (e.g., children and elderly readers) and textual domains. They also underscore the need for the development of new methods of readability scoring that will better account for reading ease.

While our work examines a variety of readability scoring methods across different reader populations and reading regimes, additional studies are needed for testing the broader generality of the presented results. First, although the OneStop dataset includes articles on diverse topics, it is restricted to the newswire domain. It is further restricted to the ``intuitive'' approach to human simplification. It does not cover human simplification approaches based on structured guidelines and auxiliary metrics, nor does it cover automated simplification. Furthermore, our study focuses on English, and the generalizability of the results to additional languages is currently an open question. Importantly, our data is limited to adult participants, and there is a need for the collection and analysis of eye tracking data from additional populations, for whom readability assessment is of central interest. Such groups include children, who are the target population of many current readability applications, as well as the elderly, and participants with reading and cognitive impairments.

Finally, \citet{kintsch-vipod} asked ``Who cares about readability?''. Their answer, based on the analysis of publications from 1941 to 1973, was ``Psychologists don't, not anymore''. Over the past decades, the psychology of reading and psycholinguistics have greatly advanced our understanding of reading processes and their behavioral traces. Although much of this research is highly relevant to readability assessment and to applications of readability in education and NLP, it has not sufficiently impacted these areas. Our study highlights some of the potential benefits of increasing the interaction of psycholinguistics with the study of readability and the application of psycholinguistic methods and insights in NLP and in applied and educational settings. 

\section*{Ethical Considerations}

The datasets OneStop \citep{onestop2025} and OneStopL2 used in this study were collected under institutional IRB protocols 1605559077 (MIT) and 157-2022 (Technion). All the participants provided written consent prior to participating in the study. The collected data is fully anonymized. Using eye tracking data for analyses of text readability is one of the use cases for which the data was collected. 

It was previously demonstrated that eye movements can be used for user identification \citep[e.g.,][]{bednarik2005eye,jager2020deep}. We do not perform user identification in this study, and emphasize the importance of not storing information that could enable participant identification in future research and applications that integrate eye movements with text readability and text simplification. 

\section*{Data and Code Availability}
OneStop \citep{onestop2025} is publicly available at \href{https://github.com/lacclab/OneStop-Eye-Movements}{https://github.com/lacclab/OneStop-Eye-Movements}. OneStopL2 will be made publicly available at \href{https://github.com/lacclab/OneStop-Eye-Movements-L2}{https://github.com/lacclab/OneStop-Eye-Movements-L2}. 
The materials and code for this paper are publicly available at 
\href{https://github.com/lacclab/Readability-Evaluation-Using-Reading-Ease}{https://github.com/lacclab/Readability-Evaluation-Using-Reading-Ease}.

\section*{Acknowledgments}
This work was supported by ISF grant 1499/22.

\bibliographystyle{compling}
\bibliography{references}

\appendix
\section*{Appendix}
\renewcommand{\thefigure}{A\arabic{figure}}
\renewcommand{\thetable}{A\arabic{table}}
\section{Textual Corpus Statistics}
\label{sec:additional-datasets}

\subsection{General Statistics}
\label{sec:additional-datasets-app}

\Cref{table:text-stats-app} presents the text statistics reported in Table~1 for OneStop, along with two additional publicly available corpora, the Parallel Wikipedia Simplification (PWKP) corpus \citep{zhu-etal-2010-monolingual} and the CommonLit Ease of Readability (CLEAR) corpus \citep{crossley2021commonlit}. These corpora are commonly used in the simplification and readability literature, and comprise texts from different domains than OneStop. 
PWKP is a corpus of comparable complex and simple sentences extracted from English Wikipedia and Simple English Wikipedia, respectively. CLEAR  is a corpus of excerpts paired with readability scores derived from manual pairwise text readability comparisons. CLEAR texts are leveled for pupils in 3rd--12th grades, and cover the informational and literary genres. The texts are drawn from the CommonLit database, Wikipedia, Project Gutenberg and open digital libraries.

Similarly to OneStop, on average, simplified sentences in PWKP are shorter and contain shorter and more frequent words than the corresponding original sentences. However, in contrast to OneStop, the average word surprisal in Simple English PWKP sentences is higher than in the English version. This outcome may be related to the finding of \citet{xu-etal-2015-problems} that about half of the Simple English sentences in PWKP are either not correctly aligned or not simpler compared to their English counterparts. While the sentences in CLEAR are comparable in length to those in OneStop, the words they contain are, on average, shorter, more frequent, and with lower surprisal than the simplified version of OneStop. The lower average complexity of CLEAR compared to OneStop is likely related to the inclusion of texts in CLEAR that are suitable for children as young as 3rd grade.

\begin{table}[ht]
\caption{Corpus statistics for the textual materials of OneStop \citep{starc2020}, PWKP \citep{zhu-etal-2010-monolingual} and CLEAR \citep{crossley2021commonlit}. Means are reported with standard deviation. Word statistics are based on whitespace tokenization. Word length excludes punctuation. In CLEAR, ``paragraphs'' refers to excerpts. Sentence boundaries for CLEAR were obtained using NLTK.}
\centering
\small
\resizebox{1\columnwidth}{!}{%
\begin{tabular}{lccccc}
\toprule
 & \multicolumn{2}{c}{OneStop} & \multicolumn{2}{c}{PWKP} & CLEAR \\
Metric & Original & Simplified & Original & Simplified &  \\
\midrule
Number of paragraphs & 162 & 162 & NA & NA & 12,010 \\
Number of sentences & 936 & 931 & 108,016 & 114,924 & 45,212 \\
Words per paragraph & 119.92 ± 27.94 & 97.14 ± 23.56 & NA & NA & 67.64 ± 47.38 \\
Sentences per paragraph & 5.78 ± 2.02 & 5.75 ± 1.76 & NA & NA & 3.76 ± 3.04 \\
Sentence length (words) & 20.76 ± 10.27 & 16.90 ± 7.90 & 21.19 ± 8.78 & 16.53 ± 7.12 & 19.13 ± 8.52 \\
Word length (characters) & 4.91 ± 2.70 & 4.72 ± 2.49 & 4.94 ± 2.78 & 4.78 ± 2.65 & 4.45 ± 2.37 \\
Word frequency (Wordfreq) & 11.26 ± 5.40 & 10.98 ± 5.23 & 12.10 ± 6.17 & 11.87 ± 6.03 & 10.76 ± 4.91 \\
Word surprisal (Pythia 70M) & 5.01 ± 4.00 & 4.77 ± 3.80 & 5.91 ± 5.23 & 5.98 ± 5.11 & 4.71 ± 3.72 \\
\bottomrule
\end{tabular}}
\label{table:text-stats-app}
\end{table}

\subsection{Types of Simplification Edits}
\label{sec:additional-datasets-simp-types}

We carried out manual annotations of simplification types in OneStop sentences. The annotations follow the simplification types taxonomy and the annotation procedure of \citet{xu-etal-2015-problems}. The taxonomy comprises three edit types: paraphrasing, deletion and split, as well as an additional no edit category. Two undergraduate students independently indicated whether each edit type is present in each sentence pair. Inter-annotator agreement was high: 94.7\% for paraphrases, 90.5\% for deletions, 98.3\% for splits, and 99.2\% for no edits. Annotation disagreements were resolved by a third annotator, who is one of the authors.

\begin{figure}[!h]
    \centering
    \includegraphics[width=\linewidth]{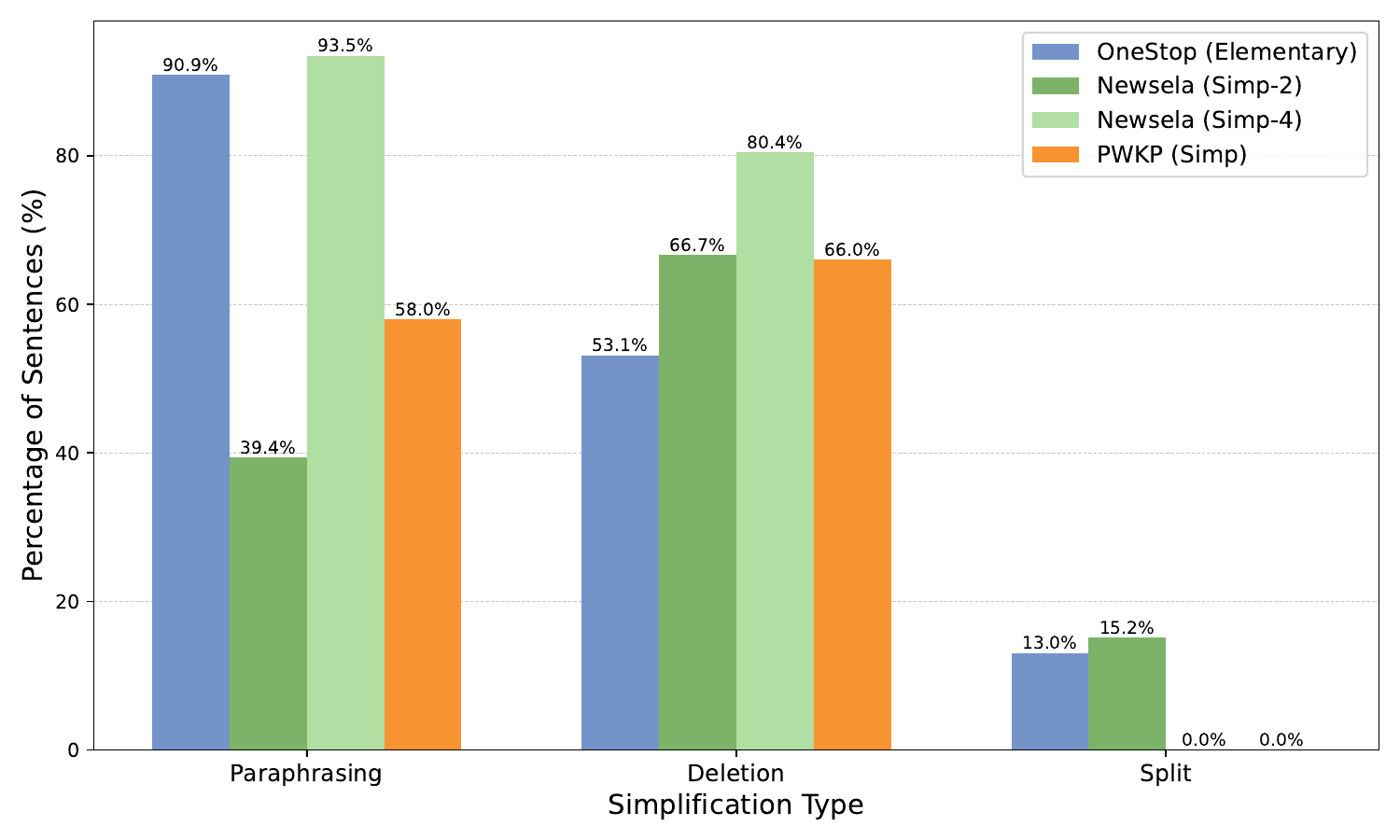}
    \caption{Simplification edit type statistics in the OneStop, Newsela, and PWKP datasets, using the edit taxonomy and edit frequency metric of \cite{xu-etal-2015-problems}. Bars represent the percentage of sentences that contain the respective simplification edit.}
    \label{fig:simplification_types_comparison}
\end{figure}

In \Cref{fig:simplification_types_comparison}, we report the prevalence of different simplification edit types in OneStop sentences along with two additional widely used simplification datasets, PWKP \citep{zhu-etal-2010-monolingual} and Newsela, analyzed in \citet{xu-etal-2015-problems}. For consistency with \citet{xu-etal-2015-problems}, and as Newsela is not publicly available, we report the same metric: percentage of sentences which include each edit type. Results for Newsela and PWKP are taken from \citet{xu-etal-2015-problems}. For Newsela, they randomly sampled 50 sentence pairs from ``Original-Simp2'' (the second simplification level) and 50 pairs from ``Original-Simp4'' (the most extensively simplified level), and manually annotated them with edit types. For PWKP, they annotated a sample of 200 aligned sentence pairs between the English Wikipedia and the Simple English Wikipedia (SIMP). We present these statistics along with statistics for OneStop for the same edit types, based on our manual edit type annotations of all the original ``Advanced'' and simplified ``Elementary'' sentence pairs in OneStop. 

We find that paraphrasing is more common in OneStop than in PWKP and Newsela Simp-2, and is comparable to the simplest Newsela version, Simp-4. Deletion, on the other hand, is less frequent in OneStop compared to the other two datasets. Sentence splitting in OneStop, which is absent from PWKP and Newsela Simp-4, is comparable to Newsela Simp-2.

\section{Statistical Testing}

\subsection{Confidence Intervals for Correlations}
In all the correlation analysis figures, we include 95\% confidence intervals for the correlation coefficients using a bootstrap over textual units (200 resamples with replacement).

\subsection{Pairwise Statistical Comparisons of Correlation Coefficients}
\label{sec:pairwise_statistical_comparisons}

To evaluate whether two correlations obtained from the same set of textual units differ significantly, we use the \textit{test for overlapping correlations based on dependent groups} implemented in the \texttt{cocor.dep.groups.overlap} function of the \texttt{cocor} package \citep{diedenhofen2015cocor}. 
The description of the test in the official documentation of \texttt{cocor}:

\begin{quote}
Performs a test of significance for the difference between two correlations based on dependent groups (e.g., the same group). The two correlations are overlapping, i.e., they have one variable in common. The comparison is made between $r_{jk}$ and $r_{jh}$. The function tests whether the correlations between $j$ and $k$ ($r_{jk}$) and between $j$ and $h$ ($r_{jh}$) differ in magnitude.
\end{quote}

Following this definition, we denote:
\begin{itemize}
    \item $r_{jk}$: correlation between variables $j$ and $k$
    \item $r_{jh}$: correlation between variables $j$ and $h$
    \item $r_{hk}$: intercorrelation between $h$ and $k$
    \item $n$: number of paired observations (texts)
\end{itemize}

We employ Steiger's (1980) two-sided test \citep{steiger1980tests}, which provides a modification of Dunn and Clark's $z$ test \citep{dunn1969correlation}. 
In all analyses, the unit of observation is the \textbf{textual unit}, with $n=162$ for paragraph-level and $n=790$ for sentence-level analyses. 

\subsubsection{Comparison between readability measures}
\label{sec:compare_readability}
In Appendix \ref{sec:main_analysis_pairwise_stat_comparisons}, which presents pairwise comparisons for the main analysis, we assess whether two readability measures ${M_1}$ and ${M_2}$ differ in how strongly their scores correlate with observed reading-ease differences across texts. 
Formally,
\begin{align*}
r_{jh} &= \text{Eval}_{M_1} = \text{Pearson}\_{\text{corr}}\big(\Delta \text{Score}_{M_1,T}, \Delta \text{ReadingEase}_{E,T} \big) \\
r_{jk} &= \text{Eval}_{M_2} = \text{Pearson}\_{\text{corr}}\big(\Delta \text{Score}_{M_2,T}, \Delta \text{ReadingEase}_{E,T} \big) \\
r_{hk} &= \text{Pearson}\_{\text{corr}}\big(\Delta \text{Score}_{M_1,T}, \Delta \text{Score}_{M_2,T}\big)
\end{align*}
where $\Delta \text{ReadingEase}_{E,T}$ is the difference in reading-ease measure $E$ (e.g., Total Fixation Duration) between the original and simplified versions of text $T$, and $\Delta \text{Score}_{M,T}$ is the difference in readability scoring method $M$ between the same pair of texts $T$.

\subsubsection{Comparison between L1 and L2 readers}
\label{sec:compare_L1-L2}
In Appendix \ref{sec:L1_L2_separate}, we use the same test to compare the correlations obtained from two dependent groups that read the same set of texts: L1 and L2 readers. 
Specifically,
\begin{align*}
r_{jh} &= \text{Eval}_{M}^{(\mathrm{L1})} = \text{Pearson}\_{\text{corr}}\big(\Delta \text{Score}_{M,T}, \Delta \text{ReadingEase}_{E,T}^{(\mathrm{L1})}\big) \\
r_{jk} &= \text{Eval}_{M}^{(\mathrm{L2})} =\text{Pearson}\_{\text{corr}}\big(\Delta \text{Score}_{M,T}, \Delta \text{ReadingEase}_{E,T}^{(\mathrm{L2})}\big) \\
r_{hk} &= \text{Pearson}\_{\text{corr}}\big(\Delta \text{ReadingEase}_{E,T}^{(\mathrm{L1})}, \Delta \text{ReadingEase}_{E,T}^{(\mathrm{L2})}\big)
\end{align*}
The correlations are considered dependent because both groups were assessed on the same texts, and the readability scoring method serves as the overlapping variable.

\subsubsection{Comparison between reading regimes}
\label{sec:compare_hunting}
Finally, in Appendix \ref{sec:information_seeking}, we test whether the relationship between text readability scores and reading ease differs across reading regimes (\textit{reading for comprehension} vs. \textit{information-seeking}). 
Here,
\begin{align*}
r_{jh} &= \text{Eval}_{M}^{(\mathrm{ReadingComp})} =\text{Pearson}\_{\text{corr}}\big(\Delta \text{Score}_{M,T}, \Delta \text{ReadingEase}_{E,T}^{(\mathrm{ReadingComp})}\big) \\
r_{jk} &= \text{Eval}_{M}^{(\mathrm{InfoSeek})} =\text{Pearson}\_{\text{corr}}\big(\Delta \text{Score}_{M,T}, \Delta \text{ReadingEase}_{E,T}^{(\mathrm{InfoSeek})}\big) \\
r_{hk} &= \text{Pearson}\_{\text{corr}}\big(\Delta \text{ReadingEase}_{E,T}^{(\mathrm{ReadingComp})}, \Delta \text{ReadingEase}_{E,T}^{(\mathrm{InfoSeek})}\big)
\end{align*}

 \clearpage

\section{Main Analysis: Pairwise Comparisons of the Reading Ease Predictivity of Readability Scoring Methods and Psycholinguistic Measures}
\label{sec:main_analysis_pairwise_stat_comparisons}

\begin{figure*}[ht]
    \centering
    \includegraphics[width=0.9\textwidth]{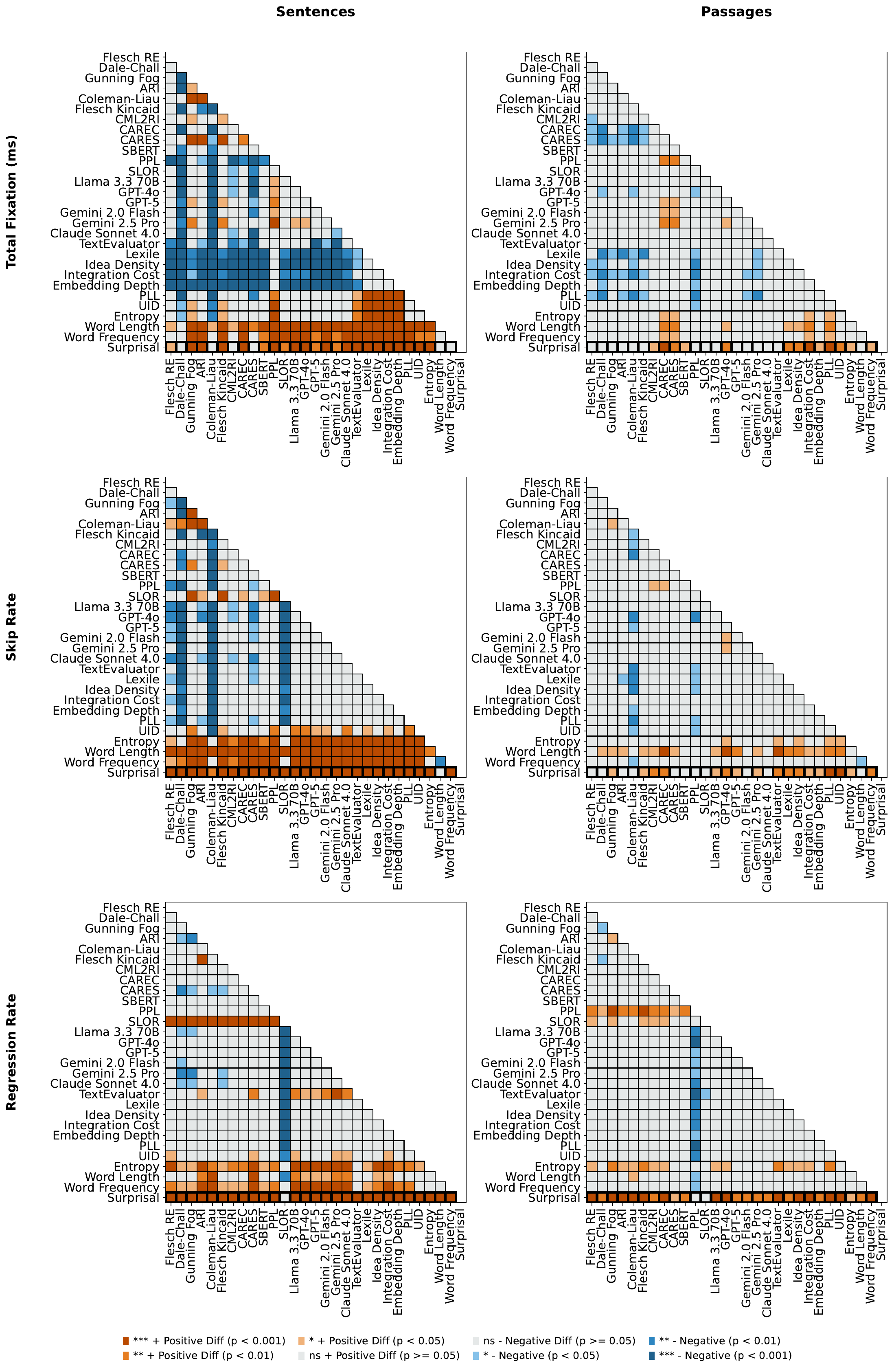}
    \caption{
\textbf{Pairwise statistical comparisons of the reading ease predictivity of readability scoring methods and psycholinguistic measures.}
Each cell $(i,j)$ compares the evaluation of readability formula $M_i$ with the evaluation of readability formula $M_j$. 
Significance is assessed using Steiger’s (1980) test for dependent overlapping correlations (see Section \ref{sec:compare_readability}). 
Cell color indicates the $p$ value of the test, with ``Positive'' and ``Negative'' denoting higher or lower correlations for the row measure relative to the column measure, respectively.
}
\label{fig:RT_perm_test_grid_RTxSenPar_boot_FirstReading}
\end{figure*}

In \Cref{fig:RT_perm_test_grid_RTxSenPar_boot_FirstReading},
each cell $(i,j)$ compares the evaluation of readability measure $M_i$ with the evaluation of readability measure $M_j$. 
Specifically, the cell is colored according to the statistical significance of comparing
$\text{Eval}_{M_i} = \text{Pearson}\_{\text{corr}}\big(\Delta \text{Score}_{M_i,T}, \Delta \text{ReadingEase}_{E,T} \big)$
with  
$\text{Eval}_{M_j} = \text{Pearson}\_{\text{corr}}\big(\Delta \text{Score}_{M_j,T}, \Delta \text{ReadingEase}_{E,T} \big)$, where $\Delta \text{ReadingEase}_{E,T}$ denotes the difference in the reading ease measure $E$ averaged across participants, between pairs of original and simplified versions of text $T$, and $\Delta \text{Score}_{M,T}$ is the corresponding difference of readability scores from method $M$. 

\clearpage

\section{Correlations between Different Readability Scoring Methods and Psycholinguistic Measures}
\label{sec:readability_measures_correlations}

\begin{figure*}[ht]
    \centering
    \includegraphics[width=1\textwidth]{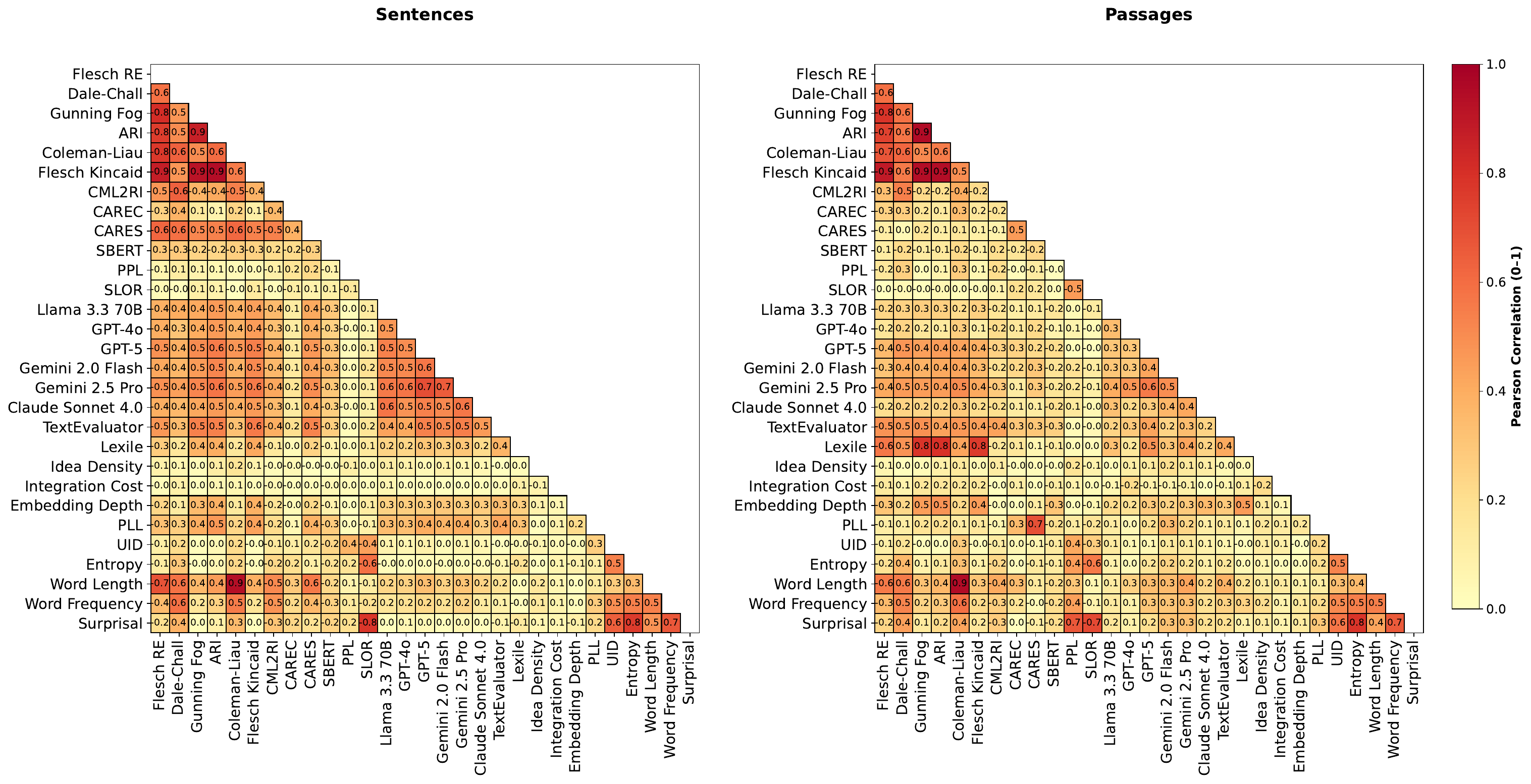}
    \caption{
    \textbf{Pairwise Pearson correlation between different readability scoring methods and psycholinguistic measures}. 
    Each cell $(i,j)$ is colored by the Pearson $r$ correlation between the readability score differences $\Delta\text{Score}_{M_i,T}$ and $\Delta\text{Score}_{M_j,T}$  produced by methods $M_i$ and $M_j$, 
    measured across all the textual items $T$ in OneStopL1\&L2. 
    }
    \label{fig:all_readability_measures_correlations}
\end{figure*}

\clearpage

\section{Analysis of Generality: Results across Different Reader Groups: L1 and L2}
\label{sec:L1_L2_separate}

\begin{figure*}[ht]
    \centering
    \includegraphics[width=1\textwidth]{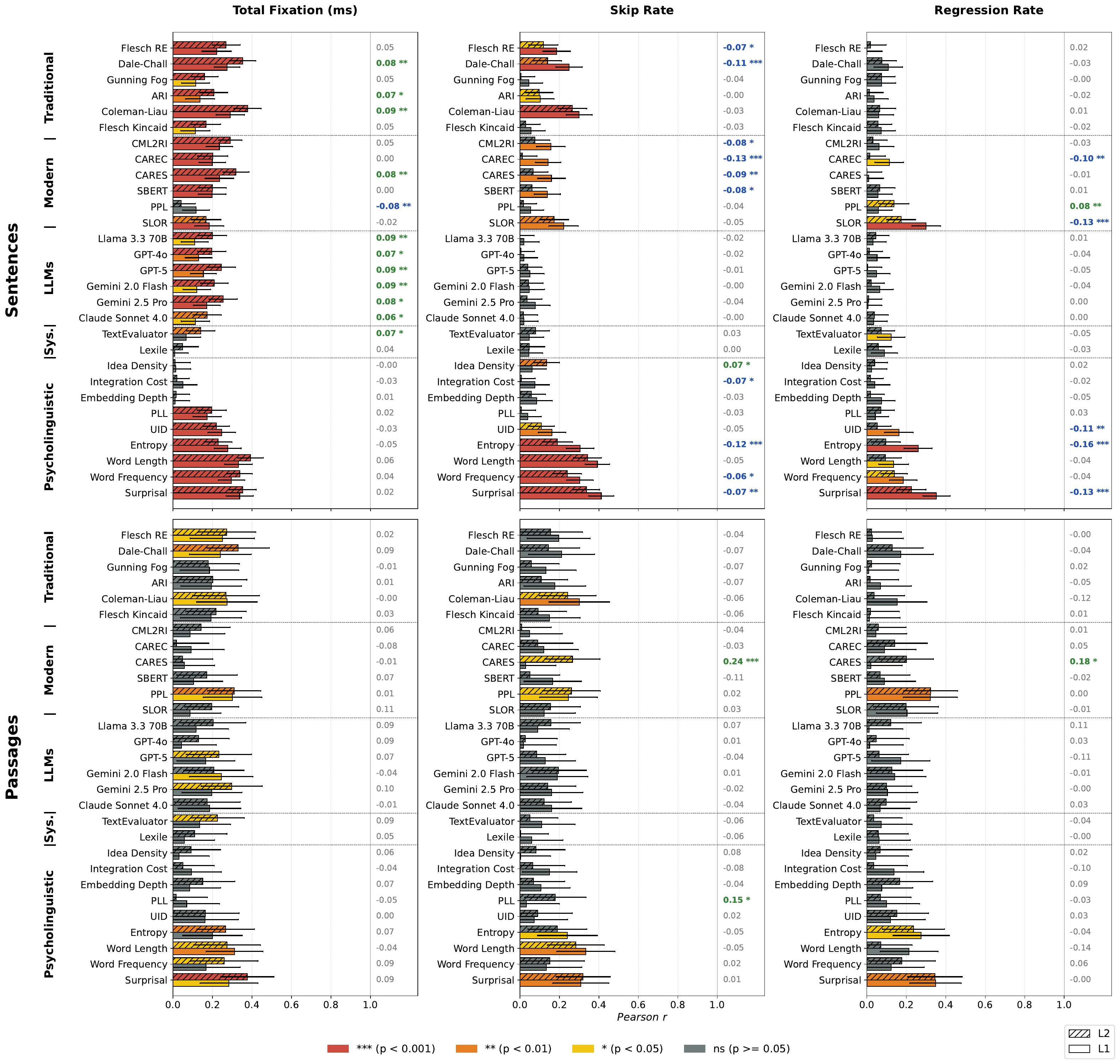}
    \caption{
Reading ease predictivity of readability scoring methods and psycholinguistic measures for \textbf{L1 and L2 speakers}.
Each pair of bars shows the Pearson correlation $r$ for \textit{L2} readers (top, striped) and \textit{L1} readers (bottom). To the right of each pair, we display the difference in correlations (L2-L1) together with its statistical significance, assessed using a Steiger correlation test (see Section \ref{sec:compare_L1-L2}). Significant differences are colored in  \textbf{\textcolor{mygreen}{green}} if 
\textbf{\textcolor{mygreen}{\textit{L2} > \textit{L1}}} and in \textbf{\textcolor{myblue}{blue}} if \textbf{\textcolor{myblue}{\textit{L1} > \textit{L2}}}. Error bars represent 95\% confidence intervals. Bar colors indicate the statistical significance level of the correlation.
}
\label{fig:SM_RT_main_RTxSenPar_pearson_corr_boot_L1_next_to_L2_FirstReading}
\end{figure*}

\Cref{fig:SM_RT_main_RTxSenPar_pearson_corr_boot_L1_next_to_L2_FirstReading} presents the main analysis  separately for English L1 and English L2 speakers.

\clearpage

\section{Analysis of Generality: Results for Different Reading Regimes}
\label{sec:information_seeking}

\begin{figure*}[ht]
    \centering
    \includegraphics[width=1\textwidth]{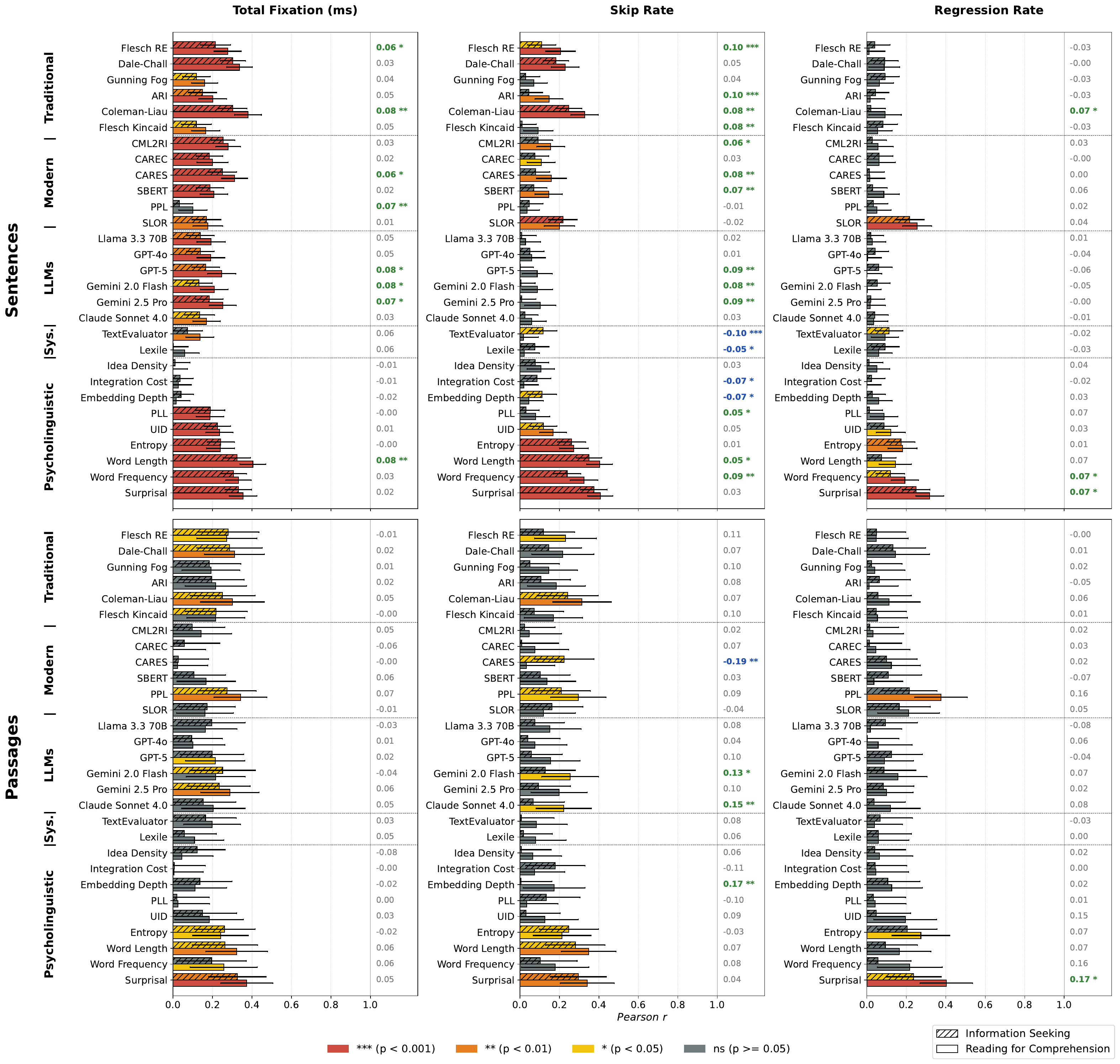}
        \caption{
Reading ease predictivity of readability scoring methods and psycholinguistic measures for \textbf{reading for comprehension} and \textbf{information seeking} reading regimes.
Each pair of bars shows the Pearson correlation $r$ for 
\textit{information seeking} readers (top, striped) and  \textit{reading for comprehension} readers (bottom). To the right of each pair, we display the difference in correlations (reading for comprehension $-$ information seeking) together with its statistical significance, assessed using a Steiger correlation test (see Section \ref{sec:compare_hunting}). Significant differences are colored in 
\textbf{\textcolor{mygreen}{green}} if
\textbf{\textcolor{mygreen}{ \textit{reading for comprehension} > \textit{information seeking}}} and in
\textbf{\textcolor{myblue}{blue}} if
\textbf{\textcolor{myblue}{\textit{information seeking} > \textit{reading for comprehension}}}. Error bars represent 95\% confidence intervals. Bar colors indicate the statistical significance level of the correlation.
}
\label{fig:SM_RT_main_RTxSenPar_pearson_corr_boot_Hunting0_next_to_Gathering0}
\end{figure*}

\Cref{fig:SM_RT_main_RTxSenPar_pearson_corr_boot_Hunting0_next_to_Gathering0} presents the main analysis separately for two between subjects reading regimes: reading for comprehension and information seeking (where the participants read the question before reading the paragraph). 

\clearpage

\section{Analysis of Generality: Additional Reading Ease Measures}
\label{sec:additional_RT_measures}
\begin{figure*}[ht]
    \centering
    \includegraphics[width=1\textwidth]{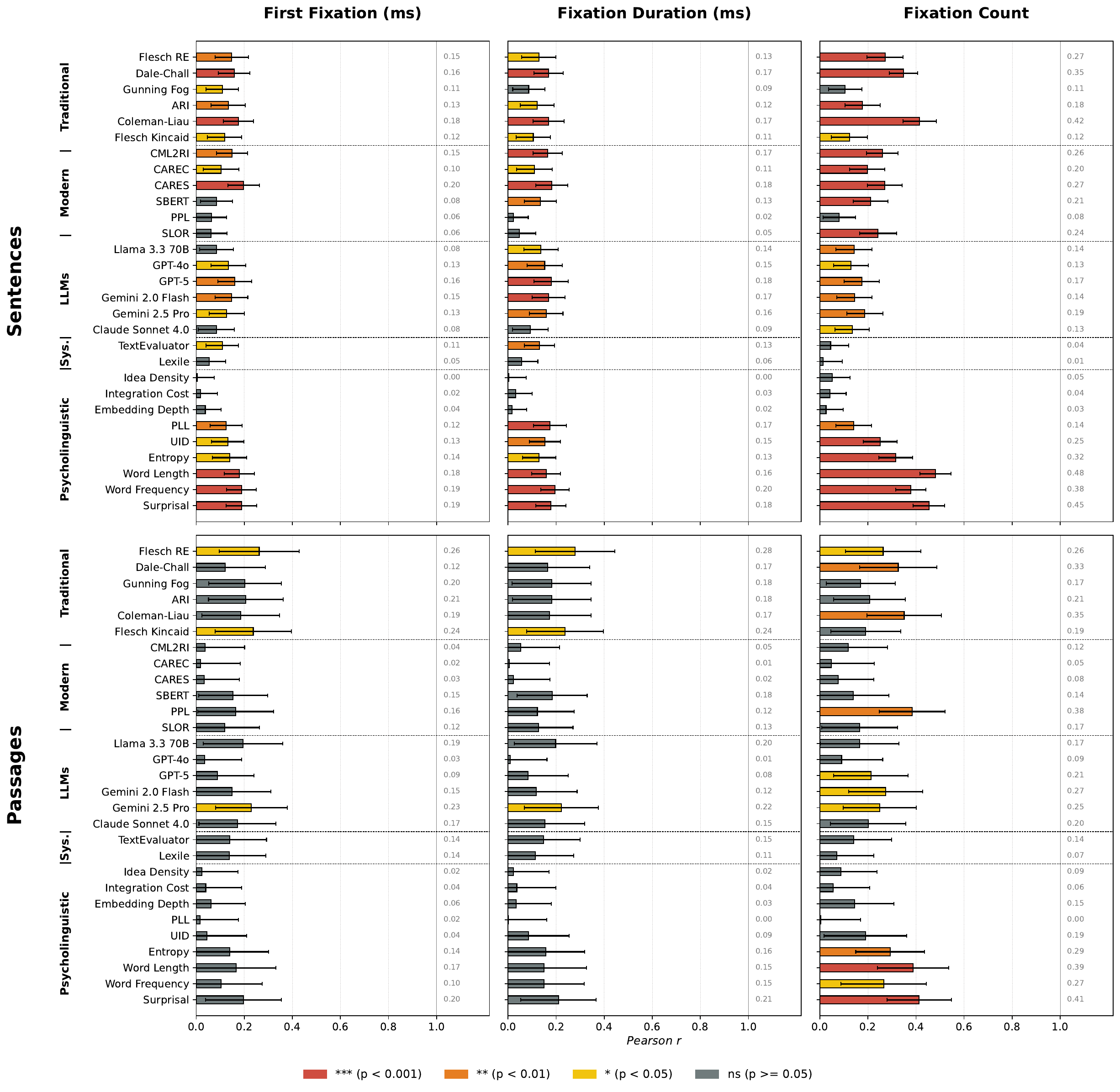}
    \caption{Reading ease predictivity of readability scoring methods and psycholinguistic measures  according to \textbf{additional eye movement measures}. Presented are Pearson correlation $r$ coefficients for \textbf{First Fixation duration (FF), Fixation Duration (FD) and Number of Fixations (NF).} 
Error bars represent 95\% confidence intervals. 
Colors represent the statistical significance level of the correlation.
}
\label{fig:SM_RT_1_RTxSenPar_pearson_corr_boot_FirstReading}
\end{figure*}

\begin{figure*}[ht]
    \centering
    \includegraphics[width=1\textwidth]{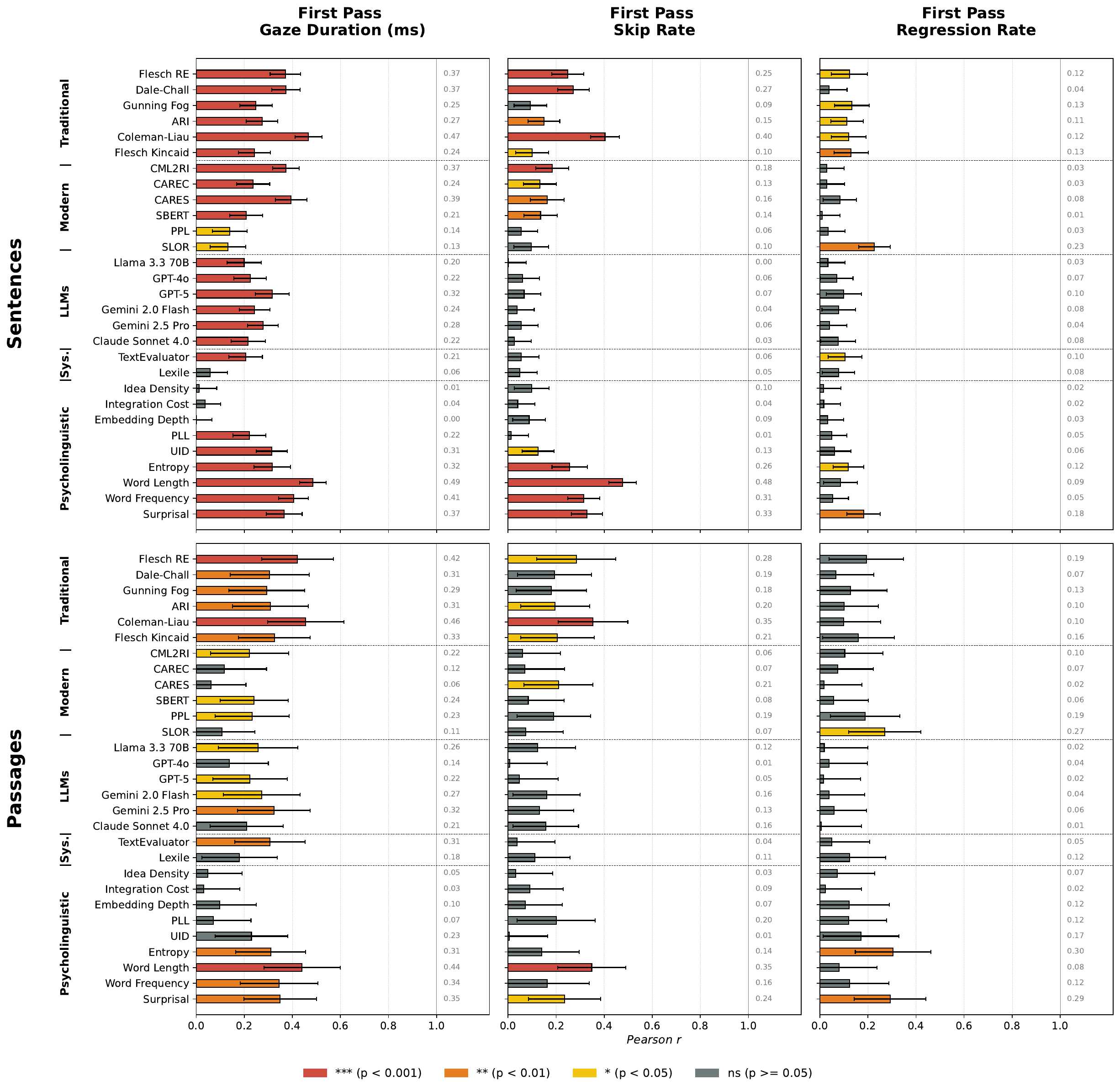}
    \caption{Reading ease predictivity of readability scoring methods and psycholinguistic measures  according to \textbf{additional eye movement measures}. Presented are Pearson correlation $r$ coefficients for \textbf{first pass Gaze Duration (fpGD), first pass Skip Rate (fpSR) and first pass Regression Rate (fpRR).} 
Error bars represent 95\% confidence intervals. 
Colors represent the statistical significance level of the correlation.
}
\label{fig:SM_RT_2_RTxSenPar_pearson_corr_boot_FirstReading}
\end{figure*}

\begin{figure*}[ht]
    \centering
    \includegraphics[width=1\textwidth]{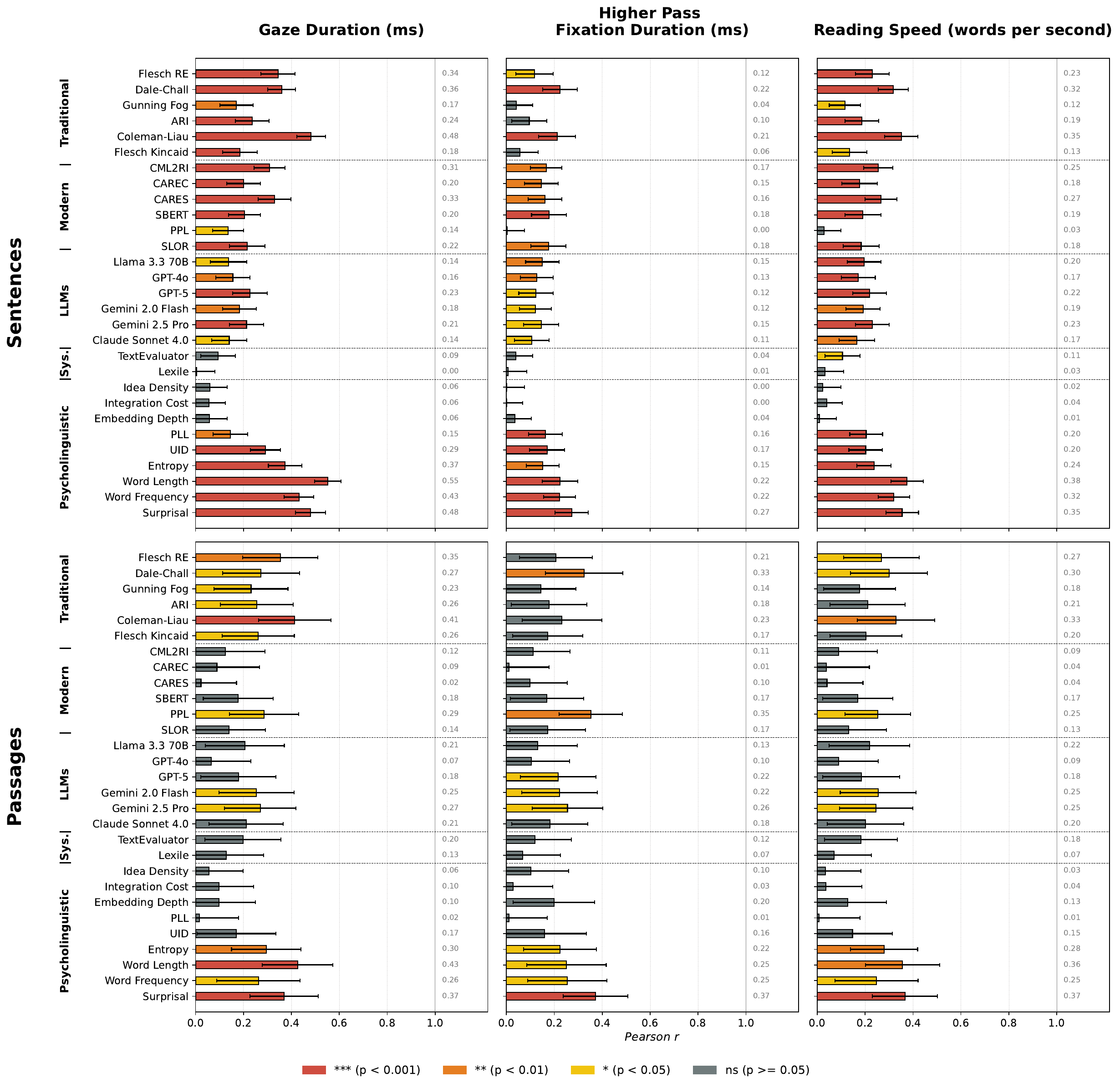}
    \caption{Reading ease predictivity of readability scoring methods and psycholinguistic measures according to \textbf{additional eye movement measures}. Presented are Pearson correlation $r$ coefficients for \textbf{Gaze Duration (GD), higher pass Fixation Duration (hpFD) and Reading Speed.} 
Error bars represent 95\% confidence intervals. 
Colors represent the statistical significance level of the correlation.
}
\label{fig:SM_RT_3_RTxSenPar_pearson_corr_boot_FirstReading}
\end{figure*}

The main analysis uses three reading ease measures, Total Fixation, Skip Rate and Regression Rate. \Cref{fig:SM_RT_1_RTxSenPar_pearson_corr_boot_FirstReading}, 
\Cref{fig:SM_RT_2_RTxSenPar_pearson_corr_boot_FirstReading}, 
and \Cref{fig:SM_RT_3_RTxSenPar_pearson_corr_boot_FirstReading} present the main analysis for additional eye tracking measures, as well as for reading speed.

\clearpage

%
\section{Analysis of Generality: Correlation Measure}
\label{sec:spearman}
\begin{figure*}[ht]
    \centering
    \includegraphics[width=1\textwidth]{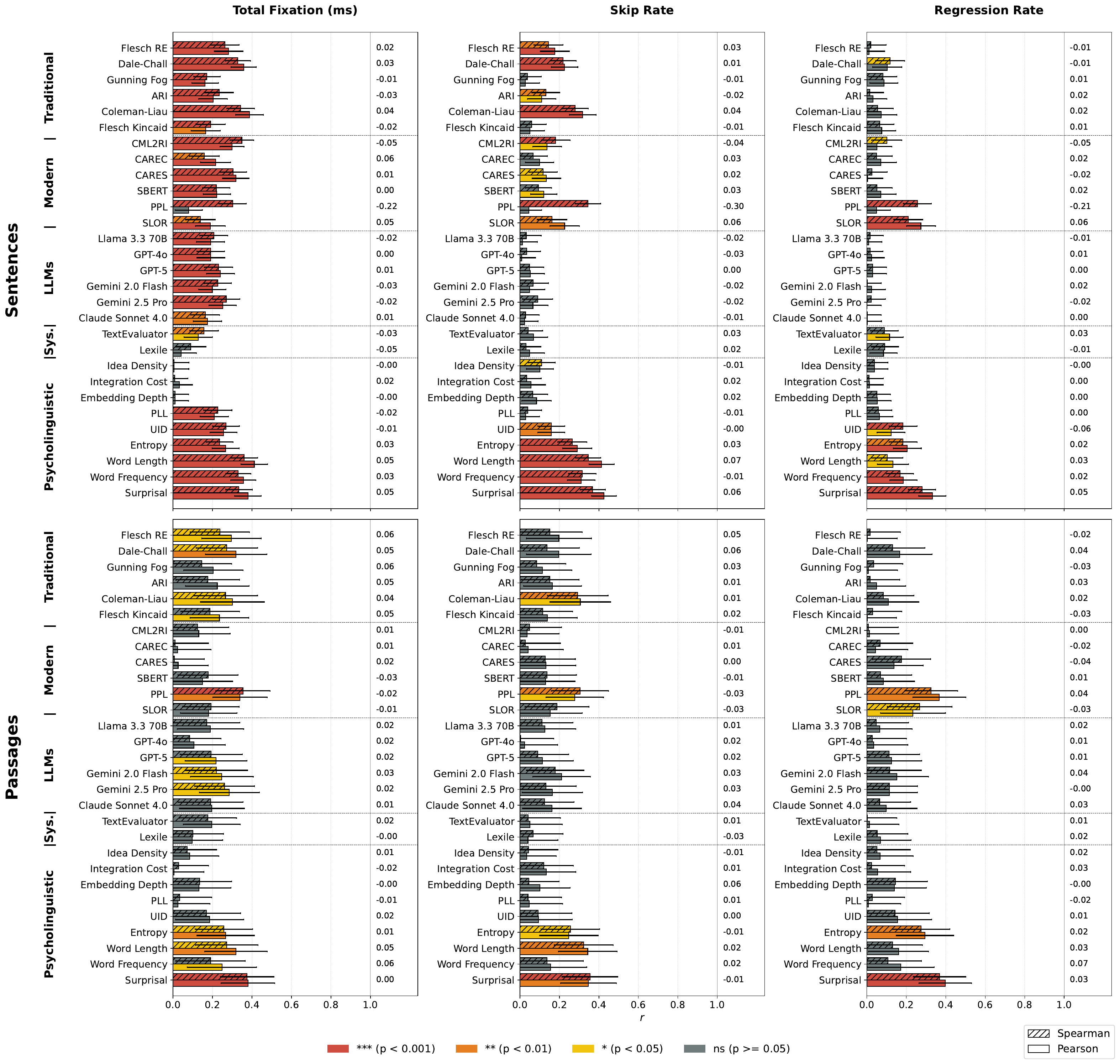}
    \caption{
Reading ease predictivity of readability scoring methods and psycholinguistic measures with \textbf{Pearson} and \textbf{Spearman} correlations. 
Each pair of bars shows the Spearman $\rho$ correlation (top, striped) and Pearson $r$ correlation (bottom). 
To the right of each pair, we display the difference between the correlation coefficients (Pearson - Spearman).
Error bars represent 95\% confidence intervals. Bar colors indicate the statistical significance level of the correlation.
}
\label{fig:SM_RT_main_RTxSenPar_pearson_next_to_spearman_corr_boot_FirstReading}
\end{figure*}

In
\Cref{fig:SM_RT_main_RTxSenPar_pearson_next_to_spearman_corr_boot_FirstReading}
we present side by side the results of the main analysis which uses Pearson correlation $r$ and the same analysis using the \textbf{Spearman  $\rho$} rank correlation coefficient. 

\clearpage

\section{Analysis of Generality: Robustness to Prompt Variants for LLMs}
\label{sec:prompt_variants}
\begin{figure*}[ht]
    \centering
    \includegraphics[width=1\textwidth]{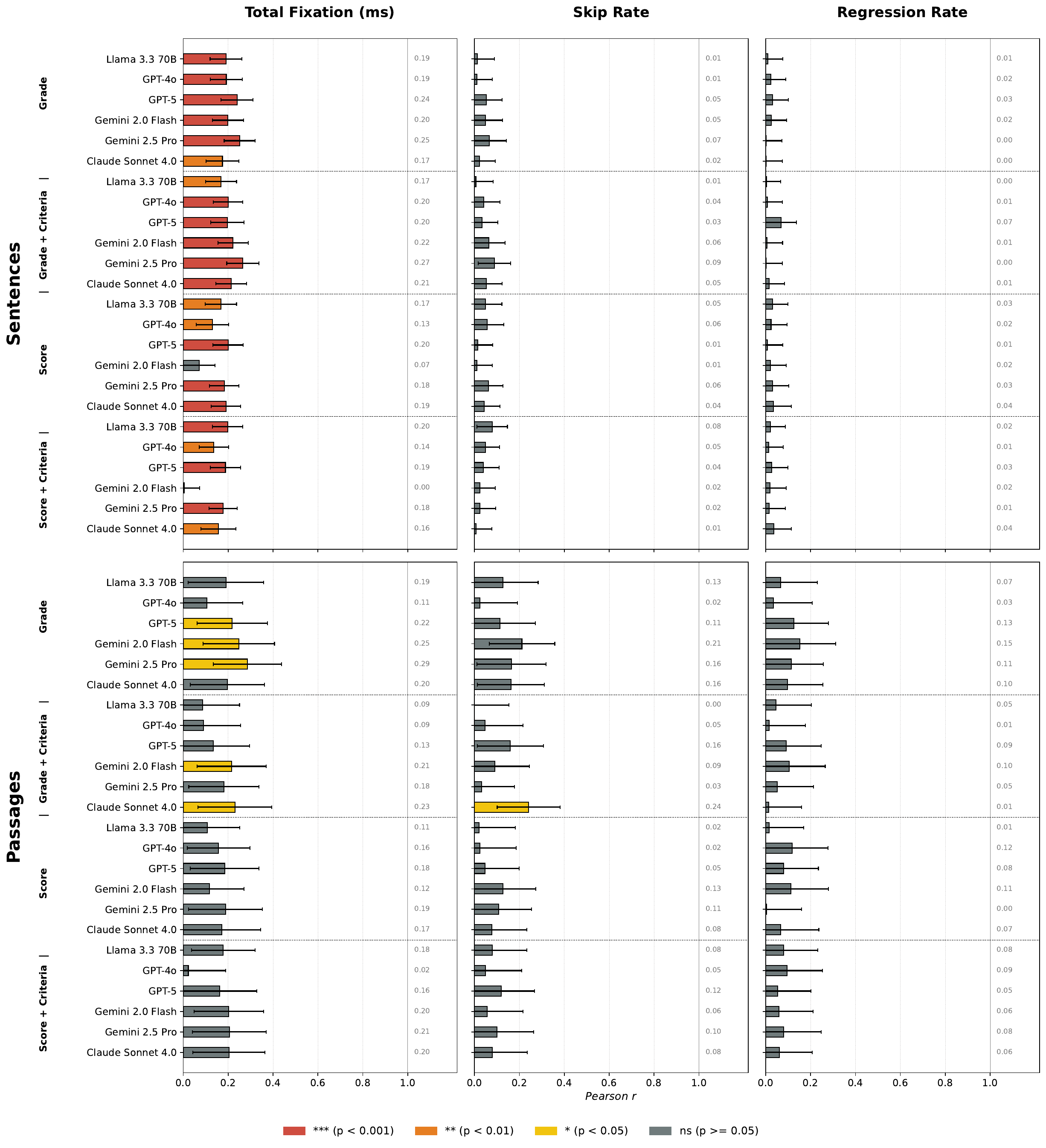}
    \caption{
\textbf{Reading ease predictivity of LLMs using different prompts.}
Presented are Pearson correlation $r$ coefficients. Error bars represent 95\% confidence intervals. 
Colors represent the statistical significance level of the correlation. 
}
\label{fig:SM_prompt_RTxSenPar_pearson_corr_FirstReading_set_main_boot}
\end{figure*}

\Cref{fig:SM_prompt_RTxSenPar_pearson_corr_FirstReading_set_main_boot} presents evaluations for LLMs using the following four prompt variants.
The first two prompts are similar to the prompts used in \cite{trott-riviere-2024-measuring}. 
The third and fourth prompts are introduced in this work and use similar wording but with a different output range of school grades, common in the readability literature. 

\begin{itemize}
    \item \textit{Score}: 
    \begin{quote}
Read the text below. 

Then, indicate the readability of the text, on a scale from 1 (very easy to read and understand) to 100 (very difficult to read and understand). 

Please answer with a single number in the range 1 to 100.

<Text>
\end{quote}
    
    \item \textit{Score with additional guidance}: 
    \begin{quote}
Read the text below. 

Then, indicate the readability of the text, on a scale from 1 (very easy to read and understand) to 100 (very difficult to read and understand). 

To determine your score, consider factors such as the complexity of sentence structure, the complexity of discourse structure, the vocabulary used, and the overall clarity of the text.

Please answer with a single number in the range 1 to 100.

<Text>
\end{quote}

    \item \textit{Grade-level}: 
    \begin{quote}
Read the text below. 

Then, indicate the readability level of the text by specifying the school grade level (1--12) for which the text would be most appropriate.

Please answer with a single number in the range 1 to 12.

<Text>
\end{quote}
    
    \item \textit{Grade-level with additional guidance}: 
    \begin{quote}
Read the text below. 

Then, indicate the readability level of the text by specifying the school grade level (1--12) for which the text would be most appropriate.

To determine your score, consider factors such as the complexity of sentence structure, the complexity of discourse structure, the vocabulary used, and the overall clarity of the text.

Please answer with a single number in the range 1 to 12.

<Text>

\end{quote}

\end{itemize}

\clearpage

\section{Results with and without Control for Text Content}
\label{sec:all_levels_results}

\begin{figure*}[ht]
    \centering
    \includegraphics[width=1\textwidth]{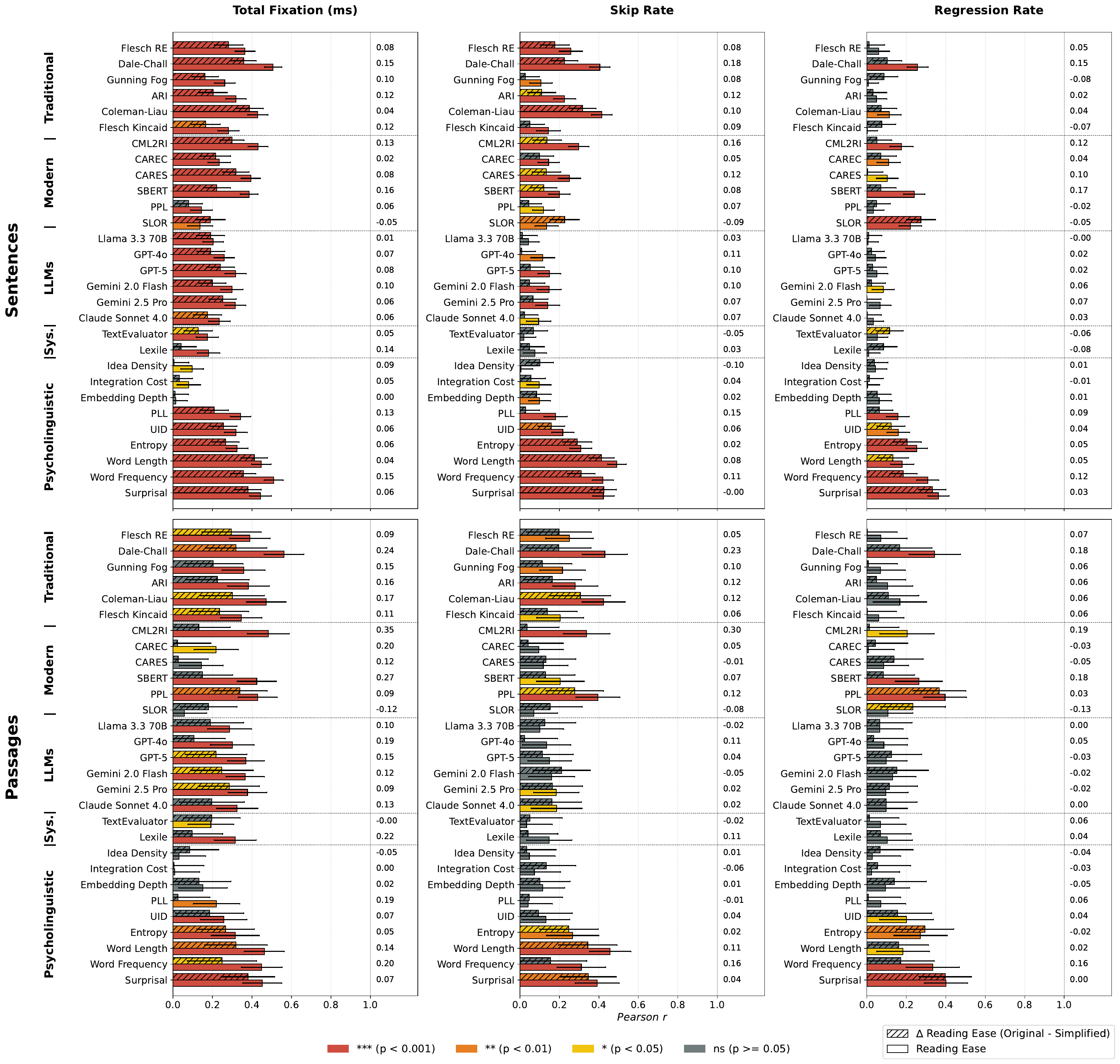}
    \caption{Reading ease predictivity of readability scoring methods and psycholinguistic measures \textbf{with and without control for text content}. Each pair of bars consists of (1) the main analysis Pearson correlation $r$ between $\Delta \text{ReadingEase}_{E,T}$ and $\Delta \text{Score}_{M,T}$ (top, striped), 
and (2) the corresponding evaluation without control for the content of the texts, where we present the Pearson correlation $r$ between $\text{ReadingEase}_{E,T}$ and $\text{Score}_{M,T}$, using all the texts in the OneStopL1\&L2 corpus (bottom). 
To the right of each pair, we display the difference in correlations ($\text{Reading Ease}$ - $\Delta \text{Reading Ease}$).
Error bars represent 95\% confidence intervals. Bar colors indicate the statistical significance level of the correlation.
}
\label{fig:SM_all_levels_pearson_corr_RE_next_to_delta_RE_boot_FirstReading}
\end{figure*}

In the main analysis, we use an evaluation methodology which controls for text content 
by regressing the differences of readability scores between original and simplified versions of the same texts against 
the corresponding differences in reading ease measures. 
Here, we compare this approach to direct regression of readability scores on reading ease measures for different texts.
The results are presented in \Cref{fig:SM_all_levels_pearson_corr_RE_next_to_delta_RE_boot_FirstReading}.

\clearpage

\section{Surprisal as a Readability Measure: Stability under Different Language Models}
\label{sec:robust_LLM}

\begin{table}[ht]
\caption{Language models used for extracting surprisal: family, model name, model size, and perplexity (PPL) measured on the OneStop Eye Movements dataset, with their corresponding Hugging Face (\url{https://huggingface.co/}) identifiers. The main analysis uses Pythia 70M.}
\centering
\small
\begin{tabular}{c|p{1.2cm}|p{3.05cm}|p{1cm}|p{1cm}|p{4.1cm}}
\toprule
\textbf{\#} & \textbf{Model Family} & \textbf{Model Name} & \textbf{Model Size} & \textbf{PPL} & \textbf{Model Identifier} \\ \midrule
1 & \multirow{7}{*}{\centering Pythia} & Pythia 70M   & 70M   & 48.99 & EleutherAI-pythia-70m     \\ 
2 &  & Pythia 160M  & 160M  & 32.87 & EleutherAI-pythia-160m    \\ 
3 &  & Pythia 410M  & 410M  & 22.61 & EleutherAI-pythia-410m    \\ 
4 &  & Pythia 1B    & 1B    & 19.43 & EleutherAI-pythia-1b      \\ 
5 &  & Pythia 1.4B  & 1.4B  & 17.70 & EleutherAI-pythia-1.4b    \\ 
6 &  & Pythia 2.8B  & 2.8B  & 15.83 & EleutherAI-pythia-2.8b    \\ 
7 &  & Pythia 6.9B  & 6.9B  & 14.63 & EleutherAI-pythia-6.9b    \\ \midrule
8 & \multirow{4}{*}{\centering GPT-2} & GPT-2 117M  & 117M  & 28.29 & gpt2                      \\ 
9 &  & GPT-2 345M  & 345M  & 21.35 & gpt2-medium               \\ 
10 &  & GPT-2 774M  & 774M  & 18.76 & gpt2-large                \\ 
11 &  & GPT-2 1558M & 1558M & 16.90 & gpt2-xl                   \\ \midrule
12 & \multirow{1}{*}{\centering GPT-J} & GPT-J 6B    & 6B    & 14.77 & EleutherAI-gpt-j-6B       \\ \midrule
13 & \multirow{3}{*}{\centering GPT-Neo} & GPT-Neo 125M & 125M & 33.20 & EleutherAI-gpt-neo-125M   \\ 
14 &  & GPT-Neo 1.3B & 1.3B & 19.58 & EleutherAI-gpt-neo-1.3B   \\ 
15 &  & GPT-Neo 2.7B & 2.7B & 17.53 & EleutherAI-gpt-neo-2.7B   \\ \midrule
16 & \multirow{2}{*}{\centering Llama-2} & Llama-2 7B   & 7B   & 9.05  & meta-llama/Llama-2-7b-hf  \\ 
17 &  & Llama-2 13B  & 13B  & 8.40  & meta-llama/Llama-2-13b-hf \\ \midrule
18 & \multirow{4}{*}{\centering OPT} & OPT 350M     & 350M & 25.54 & facebook/opt-350m         \\ 
19 &  & OPT 1.3B     & 1.3B & 18.32 & facebook/opt-1.3b         \\ 
20 &  & OPT 2.7B     & 2.7B & 16.74 & facebook/opt-2.7b         \\ 
21 &  & OPT 6.7B     & 6.7B & 15.08 & facebook/opt-6.7b         \\ \midrule
22 & \multirow{2}{*}{\centering Mistral} & Mistral-v0.1 7B & 7B & 9.21  & mistralai/Mistral-7B-v0.1 \\ 
23 &  & Mistral-v0.3 7B & 7B & 9.31  & mistralai/Mistral-7B-v0.3 \\ \midrule
24 & \multirow{3}{*}{\centering Gemma} & Gemma 7B           & 7B  & 11.55 & google/gemma-7b          \\ 
25 &  & Gemma-2 9B         & 9B  & 12.08 & google/gemma-2-9b        \\ 
26 &  & Recurrent-Gemma 9B & 9B  & 11.77 & google/recurrentgemma-9b \\ \midrule
27 & \multirow{2}{*}{\centering RWKV} & RWKV-4 169M & 169M & 28.06 & RWKV/rwkv-4-169m-pile     \\ 
28 &  & RWKV-4 430M & 430M & 21.54 & RWKV/rwkv-4-430m-pile     \\ \midrule
29 & \multirow{4}{*}{\centering Mamba} & Mamba 370M  & 370M & 19.92 & state-spaces-mamba-370m-hf \\ 
30 &  & Mamba 790M  & 790M & 17.35 & state-spaces-mamba-790m-hf \\ 
31 &  & Mamba 1.4B  & 1.4B & 15.92 & state-spaces-mamba-1.4b-hf \\ 
32 &  & Mamba 2.8B  & 2.8B & 14.42 & state-spaces-mamba-2.8b-hf \\ \bottomrule
\end{tabular}
\label{tab:models_table}
\end{table}

\begin{figure*}[ht]
    \centering
    \includegraphics[width=1\textwidth]{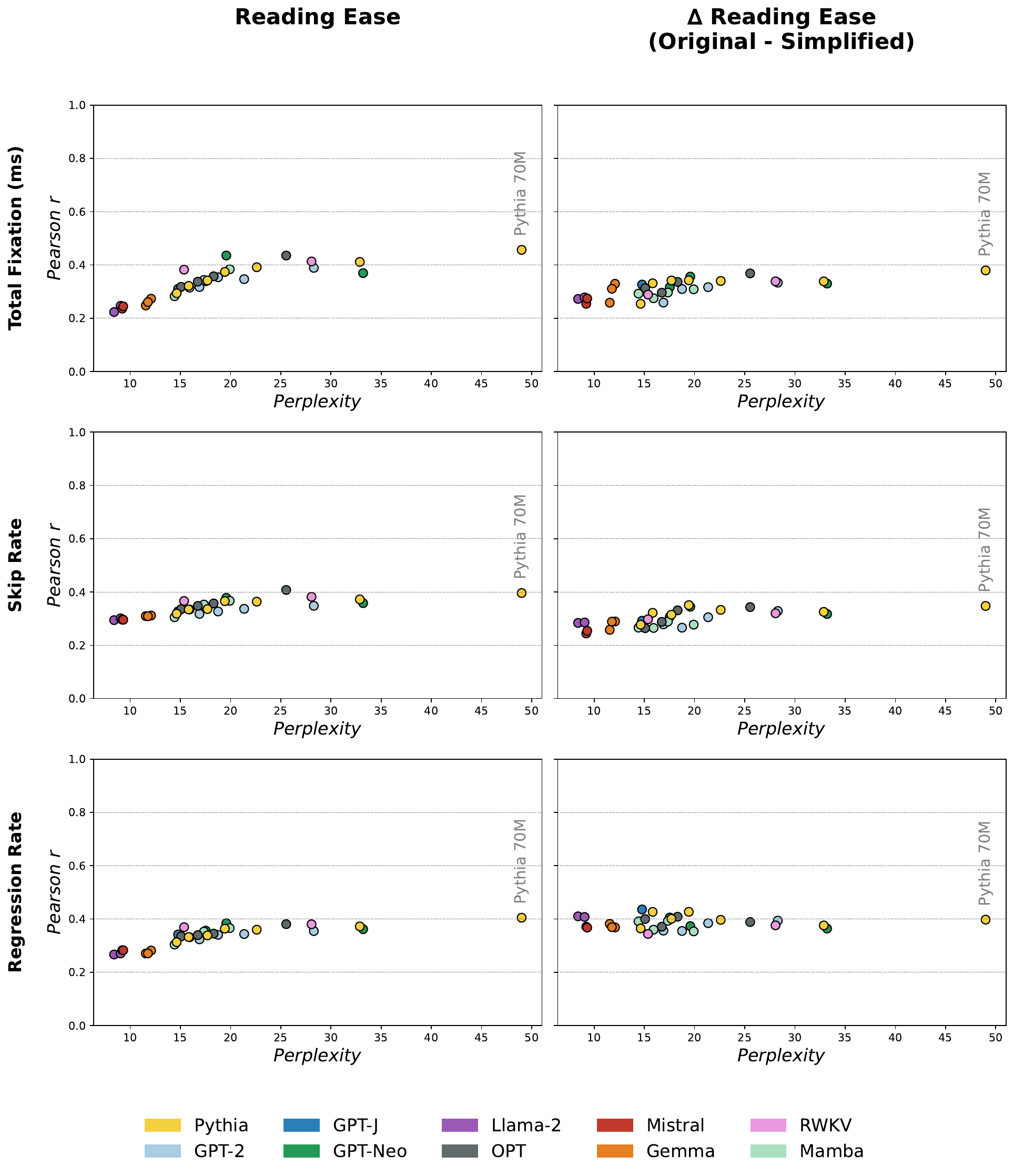}
    \caption{
\textbf{Surprisal: robustness of reading ease predictivity to the choice of language model.} Analysis at the \textbf{passage} level. 
Left: direct prediction of reading ease without text content control,
$\text{Eval}_{\text{Surprisal,LM}_i} = \text{Pearson}_{\text{corr}}\big( \text{Surprisal}_{\text{LM}_i,T}, \text{ReadingEase}_{E,T} \big)$ where $\text{Surprisal}$ are mean surprisal values per passage according to language model $\text{LM}_i$. Right: our proposed evaluation with text content control, 
 $\text{Eval}_{\text{Surprisal,LM}_i} = \text{Pearson}_{\text{corr}}\big(\Delta \text{Surprisal}_{\text{LM}_i,T}, \Delta \text{ReadingEase}_{E,T} \big)$.
Colors represent the model family. Model sizes range from 70 million to 13 billion parameters.
The main analysis uses Pythia 70M.}
\label{fig:RTxLevel_pearson_corr_by_perplexity_paragraph_FirstReading}
\end{figure*}

\clearpage

\clearpage
\section{Traditional Readability Formulas}
\label{sec:app-readability_measures}
\begin{table}[ht]
\caption{Traditional Readability Formulas}
\centering
\small
\begin{tabular}{|p{3.5cm}|p{5cm}|p{4.5cm}|}
\hline
\textbf{Name} & \textbf{Formula} & \textbf{Meaning}\\
\hline
Flesch Reading Ease \cite{flesch1948new}& 
$206.836 - 84.6 \times \frac{\text{total syllables}}{\text{total words}} - 1.015 \times \frac{\text{total words}}{\text{total sentences}}$ & 
Inversely proportional to the grade level in which 50\% of students achieved 75\% on material from the McCall-Crabbs Standard Test Lessons in Reading.\\

\hline
Dale-Chall Score \cite{dale1948formula} & 
$0.1579 \times \frac{\text{difficult words}}{\text{words}} \times 100 + 0.0496 \times \frac{\text{words}}{\text{sentences}} + 3.6365$ & 
Grade level where 50\% of the students score at least 50\% on the McCall-Crabbs Standard Test Lessons in Reading.\\

\hline
Gunning Fog Index \cite{gunning1952technique} & 
$0.4 \times \left( \frac{\text{words}}{\text{sentences}} + 
100 \times \frac{\text{complex words}}{\text{words}} \right)$ & 
Grade level, representing the estimated years of formal education required to understand the text on first reading.\\

\hline
ARI \cite{ari-smith1967} & 
$4.71 \times \frac{\text{characters}}{\text{words}} + 0.5 \times \frac{\text{words}}{\text{sentences}} $
$- 21.43$ & 
Grade level where 50\% of subjects scored at least 35\% on a cloze test. 
\\


\hline
Coleman-Liau Index \cite{coleman1975computer} & 
$0.0588 \times L - 0.296 \times S - 15.8$ \newline 
where:
\newline 
$\begin{array}{l}
L = \text{average letters} \\  \text{per 100 words} \\
S = \text{average sentences} \\  \text{per 100 words}
\end{array}$ & 
Grade level, scaled according to the expected performance of a college undergraduate in a cloze test. \\

\hline
Flesch Kincaid Grade Score \cite{flesch-kincaid1975}& 
$0.39 \times \frac{\text{total words}}{\text{total sentences}} + 11.8 \times \frac{\text{total syllables}}{\text{total words}} - 15.59$ & 
Grade level where 50\% of the subjects scored at least 35\% on a cloze test. 
\\

\hline
\end{tabular}
\label{tab:readability_measures}
\end{table}

\clearpage
\section{Tools for Computing Readability Scores}
\label{sec:implementation}

The following resources were used to implement the readability measures:
\begin{itemize}
    \item \textbf{textstat library}: version 0.7.4 \href{https://github.com/textstat/textstat}{https://github.com/textstat/textstat}, used for extracting the traditional readability measures. 
    
    \item \textbf{Automatic Readability tool for English (ARTE)}: \href{https://nlp.gsu.edu/APIdoc}{https://nlp.gsu.edu/APIdoc}
     \cite{choi2022advances}, used for extracting the modern readability measures.
    
    \item \textbf{pycpidr library}: version 0.3.0 \href{https://github.com/jrrobison1/pycpidr}{https://github.com/jrrobison1/pycpidr} \cite{brown2008automatic}, used for calculating idea density measure.
    
    \item \textbf{icy-parses}: 
    \href{https://github.com/dmhowcroft/icy-parses/commit/43f1563}{https://github.com/dmhowcroft/icy-parses}, used for calculating integration cost measure.
    
    \item \textbf{psycholing-metrics library}: version 1.1.7
    \href{https://github.com/lacclab/psycholing-metrics}{https://github.com/lacclab/psycholing-metrics} \cite{yoav_meiri_2026_20376826}, used for calculating per-word surprisal, entropy, length, and frequency.
    
    \item \textbf{ModelBlocks} incremental left-corner parser \citep{van2012connectionist} was implemented using the ModelBlocks code \href{https://github.com/modelblocks/modelblocks-release}{https://github.com/modelblocks/modelblocks-release}, used for computing embedding depth.
    
\end{itemize}

\end{document}